\documentclass[11pt]{article}

\usepackage[final]{acl}

\usepackage{times}
\usepackage{latexsym}

\usepackage[T1]{fontenc}

\usepackage[utf8]{inputenc}

\usepackage{microtype}

\usepackage{inconsolata}
\usepackage{makecell}
\usepackage{times}
\usepackage{booktabs}
\usepackage{latexsym}
\usepackage{graphicx}
\usepackage{amsthm,amsmath,amssymb,bm}
\usepackage{mathrsfs}
\usepackage{booktabs}
\usepackage{algorithm}
\usepackage{algpseudocode}
\usepackage{bbm}
\usepackage{comment}
\usepackage{multirow}
\usepackage{bbding}
\usepackage{array}
\usepackage{tcolorbox}
\usepackage{colortbl}
\usepackage{xcolor}
\usepackage{xspace}
\usepackage{enumitem}
\usepackage{threeparttable}
\usepackage{CJKutf8}
\usepackage{pifont}
\usepackage{makecell}

\newcommand{\blanksymbolfootnote}[1]{%
  \renewcommand{\thefootnote}{}
  \footnote{#1}%
  \setcounter{footnote}{0} 
  \renewcommand{\thefootnote}{\arabic{footnote}}
}

\title{IF-RewardBench: Benchmarking Judge Models for Instruction-Following Evaluation}

\author{Bosi Wen$^{1,\dagger}$ \quad Yilin Niu$^{2}$\quad Cunxiang Wang$^{2}$\quad  Xiaoying Ling$^{2}$ \\
\textbf{Ying Zhang$^{2}$\quad  Pei Ke$^{3}$\quad Hongning Wang$^{1}$ \quad Minlie Huang$^{1,\ddagger}$} \\
$^1$The Conversational Artificial Intelligence (CoAI) Group, Tsinghua University \\
$^2$Zhipu AI \quad $^3$University of Electronic Science and Technology of China   \\
\tt\small wbs23@mails.tsinghua.edu.cn, aihuang@tsinghua.edu.cn \\}

\begin{document}
\maketitle

\blanksymbolfootnote{$^\dagger$Work done when this author interned at Zhipu AI.}
\blanksymbolfootnote{$^\ddagger$Corresponding author}

\begin{abstract}
Instruction-following is a foundational capability of large language models (LLMs), with its improvement hinging on scalable and accurate feedback from judge models.
However, the reliability of current judge models in instruction-following remains underexplored due to several deficiencies of existing meta-evaluation benchmarks, 
such as their insufficient data coverage and oversimplified pairwise evaluation paradigms that misalign with model optimization scenarios. 
To this end, we propose IF-RewardBench, a comprehensive meta-evaluation benchmark for instruction-following that covers diverse instruction and constraint types. 
For each instruction, we construct a preference graph containing all pairwise preferences among multiple responses based on instruction-following quality.
This design enables a listwise evaluation paradigm that assesses the capabilities of judge models to rank multiple responses, which is essential in guiding model alignment.
Extensive experiments on IF-RewardBench reveal significant deficiencies in current judge models and demonstrate that our benchmark achieves a stronger positive correlation with downstream task performance compared to existing benchmarks. 
Our codes and data are available at \url{https://github.com/thu-coai/IF-RewardBench}.

\end{abstract}

\section{Introduction}
Large language models (LLMs) have exhibited remarkable capabilities across a broad range of NLP tasks \cite{zhao2023survey}.
Among these, instruction-following is a foundational requirement for practical LLM applications \cite{ouyang2022training}.
In real-world deployments, most tasks are formulated as an instruction-following paradigm, where human instructions specify task requirements and impose corresponding constraints on model outputs \cite{jiang-etal-2024-followbench}.
Therefore, faithfully following instructions is critical to ensure the reliability of LLMs and enable their generalization to novel and complex tasks \cite{huang2024survey}.

Improving this capability of LLMs hinges on an accurate evaluation of whether model responses follow each constraint within instructions. 
Rigorous evaluation not only enables reliable progress measurement and failure modes diagnosis \cite{wen2024benchmarking, qin2025sysbench}, but also provides the reward signals essential for steering model alignment \cite{peng-etal-2025-verif, qin2025incentivizing}. 
Recently, LLM-as-a-Judge has been widely adopted as a scalable approach for this purpose \cite{liu2025recast, he2025rubric}. 
These developments underscore the need for a systematic examination of its reliability for instruction-following evaluation.

\begin{figure}[t]
\scriptsize
    \centering
    \includegraphics[width=1.0\linewidth]{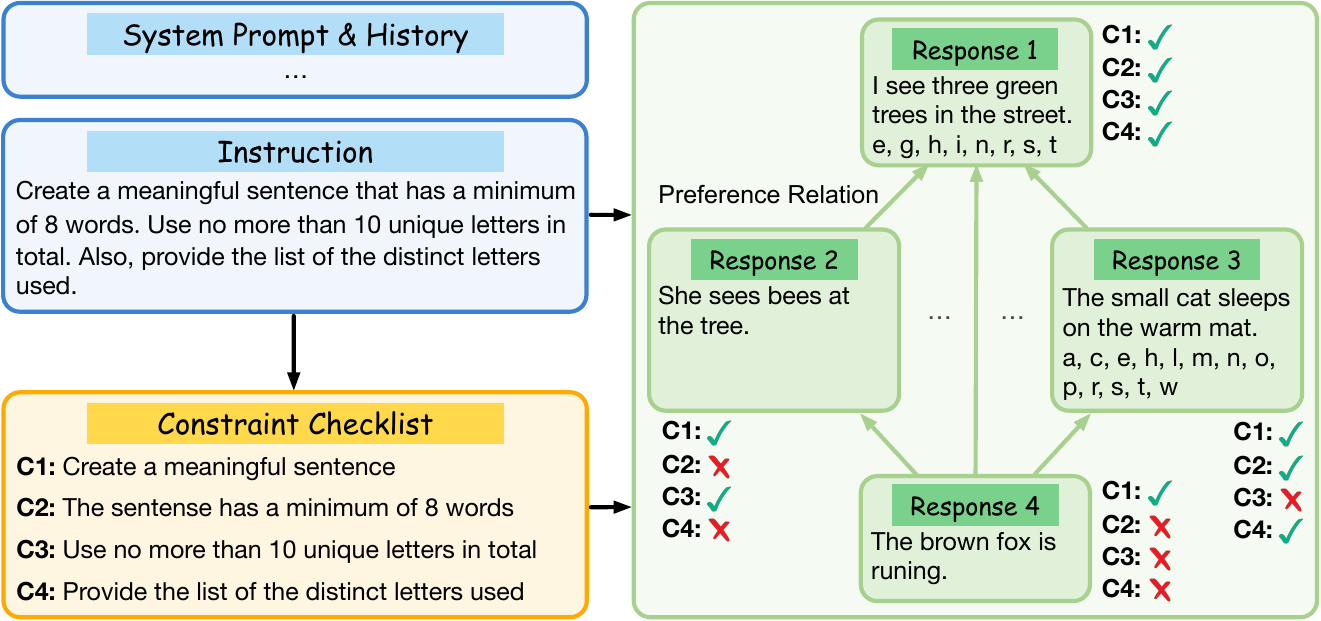}
    \caption{An example from IF-RewardBench, containing a user instruction, a constraint checklist, and multiple responses with various instruction-following quality that form a preference graph. }
    \label{fig:example}
\end{figure}

As previous benchmarks for judge models often focus on reasoning or general chat, neglecting fine-grained instruction-following  \cite{tan2025judgebench, lambert-etal-2025-rewardbench},  specialized benchmarks have emerged to fill this gap \cite{peng-etal-2025-agentic, malik2025rewardbench}.
They typically ask judge models to identify the best response from multiple candidates that follows all constraints.
Despite these efforts, they still suffer from three serious limitations:
(1) \textbf{Insufficient Data Coverage}: 
existing benchmarks predominantly focus on single-turn instruction and narrow constraint types (i.e., code-verifiable constraints and their \textit{And} composition) \cite{frick2025how}, failing to capture the heterogeneity of real-world user instructions,
which frequently involve system prompts, conversation history, and a diverse spectrum of constraints.
This discrepancy hinders a comprehensive and robust evaluation.
(2) \textbf{Oversimplified Evaluation Paradigms}: 
Realistic model optimization scenarios require judge models to precisely rank multiple responses of varying quality to derive relative reward advantages for parameter updates \cite{shao2024deepseekmath}. 
However, prevailing pairwise or Best-of-N (BoN) selection paradigms reduce evaluation to a winner-take-all decision that merely identifies the single best response.
This oversimplification ignores the intricate partial order among multiple responses and limits the assessment of the ranking capability essential for effective alignment guidance.
(3) \textbf{Unreliable Ground Truth Labels}: Many existing benchmarks solely rely on judge models or evaluation scripts to construct preference pairs without human verification, rendering them susceptible to evaluation bias \cite{zheng2023judging} and confounding factors unrelated to instruction-following.

\begin{table*}[!t]
\centering

\resizebox{\linewidth}{!}{
\begin{tabular}{@{}lccccccccc@{}}
\toprule

\multirow{2}{*}{Benchmark}  & \multicolumn{4}{c}{Dataset Size}  & Diverse  & \multicolumn{2}{c}{Instruction Type}  & \multirow{2}{*}{Annotator} & \multirow{2}{*}{Evaluation Paradigms}      \\
 \cmidrule(lr){2-5} \cmidrule(lr){7-8}
  & \#Inst. & \#Resp. & \#Rela. & \#Model & Constraint & Multi-Turn & System Prompt   & \\        
\midrule

LLMBar~\citeyearpar{zeng2024evaluating}    & 419  & 836  & 419 & 3  & \color{red}\ding{55}  & \color{red}\ding{55} & \color{red}\ding{55} & Human & Pairwise \\

InfoBench~\citeyearpar{qin-etal-2024-infobench}  & 50  & 250 & 0 & 5    & \color{orange}$\triangle$ & \color{red}\ding{55}  & \color{red}\ding{55} & Human & Pointwise \\

IFBench~\citeyearpar{peng-etal-2025-agentic} & 444  & 888 & 444 & 1 & \color{orange}$\triangle$  & \color{red}\ding{55} & \color{red}\ding{55} & GPT-4o & Pairwise \\

PPE-IF~\citeyearpar{frick2025how} & 512  & 16384 & 2560 & 4 & \color{orange}$\triangle$ & \color{red}\ding{55}  & \color{red}\ding{55} & Synthetic & Paiwise \& BoN \\

RewardBench-2-IF~\citeyearpar{malik2025rewardbench} & 160 & 640   & 480 & 5  &  \color{orange}$\triangle$ & \color{red}\ding{55}  & \color{red}\ding{55} & Synthetic & BoN \\

\midrule
\textbf{IF-RewardBench (ours)}   & \textbf{842} & \textbf{6011}  & \textbf{9145} & \textbf{16} & \color{green!60!black}\ding{51} & \color{green!60!black}\ding{51} & \color{green!60!black}\ding{51} & \textbf{Human} & \textbf{Pointwise \& Listwise} \\ 

\bottomrule
\end{tabular}
}

\caption{Comparisons of IF-RewardBench and other meta-evaluation benchmarks for instruction-following. 
The data size contains the number of instructions (\#Inst.), responses (\#Resp.), preference relations (\#Rela.), and models to generate responses (\#Model). \textcolor{orange}{$\triangle$} denotes partially satisfied. }
\label{tab:comparisons}
\end{table*}
To this end, we propose IF-RewardBench, a comprehensive benchmark to assess the capability of judge models in instruction-following evaluation, which is characterized by three key advantages:
\begin{itemize}[leftmargin=*]
\item \textbf{Comprehensive Coverage.} 
IF-RewardBench comprises 842 instructions covering 3 critical instruction types: \textit{single-turn interaction}, \textit{multi-turn interaction}, and \textit{system-prompt steerability}. 
These instructions include a diverse spectrum of constraints and their compositions. 
A total of 6,011 responses are generated by 16 different LLMs to ensure diversity.
\item \textbf{Realistic Evaluation Paradigms.} 
We introduce a novel \textit{listwise} evaluation paradigm for judge models extending from pairwise or BoN selection.
For each instruction, we collect multiple responses and annotate their adherence to each constraint within the instruction.
Based on the Pareto dominance relations derived from these annotations,  we establish all pairwise preferences among these responses to construct a preference graph, as shown in Figure \ref{fig:example}.
Judge models are required to rank these responses to best align with the underlying preferences, a setting that closely mirrors realistic model optimization scenarios.
\item \textbf{Reliability.} 
Each example in IF-RewardBench is annotated by multiple human experts and undergoes multi-step rigorous inspection, ensuring data quality and evaluation reliability.
\end{itemize}

We conduct a comprehensive evaluation of 22 popular judge models on IF-RewardBench, including state-of-the-art dedicated reward models and general LLMs. 
Our results highlight a substantial capability gap: even the leading proprietary LLM, Gemini-3-Pro, achieves only a moderate Kendall correlation of 0.609 in ranking responses,
significantly falling behind the human performance of 0.755. 
Top-tier open-source models, such as GLM-4.6 and Deepseek-V3.2, remain below or near 0.4, while all dedicated reward models fail to exceed 0.2.
In-depth analysis reveals several critical insights:
(1) Although constraint-level scoring outperforms overall pairwise comparison, the limited capacity for detecting constraint violation remains a primary performance bottleneck.
(2) Incorporating system instructions or conversation history exacerbates evaluation difficulty for overall pairwise comparison.
(3) Situation and Style constraints that involve subjectivity prove harder to verify than objective ones.
(4) Judge performance degrades with increased constraint composition complexity, number of constraints within instructions, and response quality, underscoring the growing challenge of evaluating advanced models in complex application scenarios.
Notably, IF-RewardBench presents greater difficulty than existing benchmarks and demonstrates a significantly stronger positive correlation with the downstream BoN sampling performance of judge models.
These findings establish IF-RewardBench as a valuable resource for advancing instruction-following evaluation.

\section{Related Work}

\paragraph{Instruction-Following.} 
As LLMs are increasingly applied to address complex real-world tasks, instruction-following emerges as a key determinant of their practical utility \cite{liu2023trustworthy, lou2024large}, motivating extensive efforts to evaluate and improve this capability of LLMs. 
For evaluation, numerous instruction-following benchmarks have been proposed from various perspectives. 
They often employ judge models to assess whether the model response can follow each constraint in the input instruction \cite{jiang-etal-2024-followbench, qin-etal-2024-infobench, wen2024benchmarking, qin2025sysbench, zhang-etal-2025-cfbench}. 
For improvement, a growing body of work attempts to leverage reward signals from judge models to drive instruction-following optimization via preference or reinforcement learning \cite{zhang-etal-2025-iopo, cheng2025spar, ren-etal-2025-step, liu2025recast, peng-etal-2025-verif, wen2025if}. 
Consequently, judge models serve as a linchpin in both the evaluation and optimization pipelines. 
However, their reliability in this domain remains insufficiently explored.

\paragraph{Evaluation of Judge Models.} 
Given the pivotal role of judge models in alignment, numerous benchmarks have been introduced to evaluate their capability across diverse domains, including chat \cite{lambert-etal-2025-rewardbench, zhou2025rmb}, reasoning \cite{tan2025judgebench}, and agents \cite{men-etal-2025-agent, lu2025agentrewardbench}.  
These benchmarks typically curate instruction-winner-loser triplets to assess the accuracy of judge models in making preference judgments. 
For instruction-following, IFBench \cite{peng-etal-2025-agentic} employs GPT-4o to synthesize preference pairs. 
PPE \cite{frick2025how} and RewardBench 2 \cite{malik2025rewardbench} focus on code-verifiable constraints derived from IFEval \cite{zhou2023instruction}, employing both the pairwise and BoN evaluation paradigms. 
However, existing benchmarks mainly consider single-turn instructions and a narrow set of constraints. 
The lack of human verification could diminish evaluation reliability.
And the pairwise or BoN evaluation paradigms may diverge from realistic model optimization scenarios. 
These are the limitations our work aims to address.

\section{IF-RewardBench}
\label{sec:construction}
\begin{figure*}[!t]
\scriptsize
    \centering
    \includegraphics[width=1\textwidth]{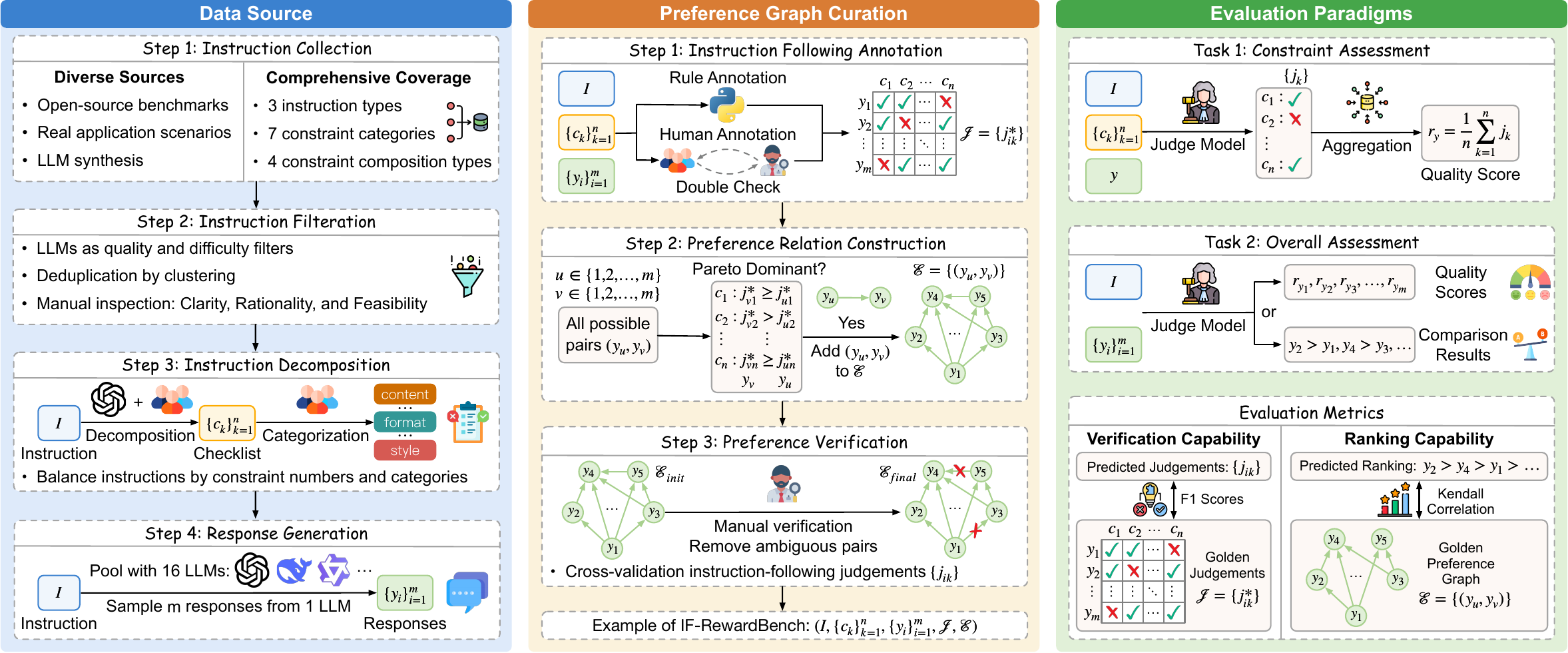}
    \caption{Overall framework of IF-RewardBench. \textit{Left}: Collect instructions and responses from diverse sources. \textit{Center}: Curate preference graphs via multi-stage annotation and verification. \textit{Right}: Assess various judge models based on different evaluation paradigms.}
    \label{fig:overview}
\end{figure*}

\subsection{Task Definition}
To conduct a comprehensive meta-evaluation of instruction-following, we first define the evaluation tasks and identify the core capabilities required for effective evaluation.
\subsubsection{Evaluation Tasks of Judge Models}
Following previous works \cite{ wen2024benchmarking, peng-etal-2025-verif, ren-etal-2025-step, malik2025rewardbench}, we consider two prevalent instruction-following evaluation tasks of judge models:

\noindent 1) \textbf{Constraint Assessment}: Given an instruction $I$ (which may contain the system prompt and conversation history), a constraint checklist $\{c_k\}^n_{k=1}$ derived for $I$, and a model response $y$, the judge model is tasked to evaluate whether $y$ follows each constraint $c_k$, providing a binary judgement $j_k \in \{0,1\}$ for each constraint.
The instruction-following quality of $y$ is then quantified as:
\begin{equation}
\label{eq:quality_reward}
\begin{aligned}
r_y = \frac{1}{n}\sum_{k=1}^nj_{k}
\end{aligned}
\end{equation}
2) \textbf{Overall Assessment}: Given an instruction $I$ and a set of model responses $\{y_i\}_{i=1}^m$, the judge model is tasked to provide quality scores or pairwise comparison results for these responses.

\subsubsection{Core Capabilities of Judge Models}
Ideally, a judge model should possess two core capabilities: 
(1) \textbf{Verification}: Accurately verify each constraint in constraint assessment, enabling granular progress tracking.
(2) \textbf{Ranking}: Ensure that the assessments for multiple responses (derived via Equation \ref{eq:quality_reward} or directly generated) faithfully reflect the partial order of their instruction-following quality, thereby guiding model alignment.
IF-RewardBench is designed to comprehensively evaluate judge models across both aspects. 

\subsection{Dataset Construction} 

In this section, we introduce the construction of IF-RewardBench, as illustrated in Figure \ref{fig:overview}.
Each example in IF-RewardBench is structured as a preference graph $(I, \{c_k\}_{k=1}^n, \{y_i\}_{i=1}^m, \mathcal{J}, \mathcal{E})$.
The set $\mathcal{J}$ contains the ground truth following judgement $j^*_{ik}$ for each response $y_{i}$ to each constraint $c_{k}$, providing the basis for assessing \textbf{Verification}.
And the set $\mathcal{E}$ contains all preference relations among these responses: a directed edge $(y_u, y_v) \in \mathcal{E}$ indicates that $y_v$ has better instruction-following quality than $y_u$, providing the basis for assessing \textbf{Ranking}. 
Unlike previous benchmarks that typically curate a single preference pair per instruction \cite{zeng2024evaluating, lambert-etal-2025-rewardbench, men-etal-2025-agent}, our preference graph captures the complex partial order among multiple responses, which better accommodates the realistic application of judge models in alignment guidance.
To construct IF-RewardBench, we first collect instructions and responses from diverse sources (§\ref{data_source}), and then utilize a multi-stage annotation and verification pipeline to curate preference graphs (§\ref{data_curation}).

\subsubsection{Data Source}
\label{data_source}
\paragraph{Instruction Collection.}
To obtain diverse and representative instructions for benchmark construction, we collect instructions from real-world application scenarios and 14 open-source instruction-following benchmarks. 
This collection covers a diverse spectrum of constraints and 3 critical instruction types: \textit{single-turn interaction}, \textit{multi-turn interaction}, and \textit{system-prompt steerability}. 
Multi-turn interaction requires following constraints carried from previous turns \cite{he2024multi, he2025rubric}, while system-prompt steerability requires following system prompts and prioritizing them over user prompts \cite{qin2025sysbench, zhang-etal-2025-iheval}.
Given the scarcity of complex constraint compositions in existing benchmarks \cite{wen2024benchmarking, qin2025incentivizing}, we further leverage LLMs to synthesize complex instructions based on seeds from real-world application scenarios, guided by a comprehensive taxonomy encompassing 7 primary constraint categories (Numerical, Format, Content, Linguistic, Style, Situation, and Action) and 4 constraint composition types (\textit{Single}, \textit{And}, \textit{Chain}, and \textit{Selection}). 
In total, we collect approximately 24.6k instructions. 
Details of the constraint taxonomy and instruction sources are in Appendix \ref{app:constraint_taxonomy} and \ref{app:instruction_source}.

\paragraph{Instruction Filtration.} 
We then implement a multi-step instruction filtration process to ensure their quality and complexity.
After heuristic length filtering, we first utilize LLMs to score instruction quality and complexity, retaining only those with high scores on both aspects.
Detailed prompts are in Appendix \ref{app:prompt_templates}. 
Then, we select a representative subset via DBSCAN \cite{ester1996density} clustering, based on the embeddings of instructions using Conan-embedding \cite{li2024conan}. 
Finally, manual inspection is conducted to eliminate three problematic cases: (1) those with unreasonable, ambiguous, or inconsistent constraints, (2) those beyond LLM capabilities (e.g., image generation), and (3) those demanding highly specialized domain knowledge. 
This process yields a final set of 3,978 instructions.

\paragraph{Instruction Decomposition.} 
For each instruction $I$, we employ LLMs to automatically decompose constraints and generate a checklist $\{c_k\}^n_{k=1}$, whose prompt is in Appendix \ref{app:prompt_templates}. 
Existing checklists from source benchmarks are directly adapted where available.
All checklists are then manually revised to correct any errors.
For each constraint, we also ask annotators to label all included constraint categories and composition types following the above taxonomy. 
Finally, we balance instructions by constraint numbers and categories, yielding 2,459 instructions with high representativeness.
\paragraph{Response Generation.}
To obtain diverse responses for our elaborately curated instructions, we utilize 16 widely used LLMs with varying capabilities to generate responses. For each instruction, all responses are generated by a single LLM, which can effectively control confounding variation unrelated to instruction-following, such as writing quality and style \cite{malik2025rewardbench}. 
Appendix \ref{app:model_list} details the LLMs used for response generation.

\subsubsection{Preference Graph Curation}
\label{data_curation}
\paragraph{Instruction-Following Annotation.}
We first obtain the golden truth following judgement $j^*_{ik}$ for each response $y_i$ to each constraint $c_k$, enabling the evaluation of verification capability and preference derivation in subsequent steps. 
For instructions with provided evaluation scripts, we employ them directly.
For other instructions, we recruit 22 college students for manual evaluation.
All of them pass the mandatory proficiency examination and annotation tutorial.
Each response is independently judged by two annotators, with a third inspector conducting spot checks. 
Any discrepancies are re-annotated by the inspector and discussed to reach a consensus.
Details of human annotation and quality control processes are in Appendix \ref{app:human_annotation}.

\paragraph{Preference Relation Construction.}
Based on the judgements, we examine all possible pairs to construct preference relations. 
Since instructions contain multiple constraints, deriving preferences from the quality score according to Equation \ref{eq:quality_reward} may induce constraint-level inconsistencies and ambiguity. 
To address this, we only retain pairs where the positive response Pareto dominates the negative one across all constraints. 
Formally, a preference relation $(y_u, y_v)$ is constructed if and only if:
\begin{equation}
\label{eq:hyp_1}
\begin{aligned}
\forall k \in \{1,2,…,n\}, j^*_{vk} \ge j^*_{uk}
\end{aligned}
\end{equation}
\begin{equation}
\label{eq:hyp_2}
\begin{aligned}
\exists  k \in \{1,2,…,n\}, j^*_{vk} > j^*_{uk}
\end{aligned}
\end{equation}
\paragraph{Preference Verification.}
To further enhance data quality, we manually review and filter the constructed preference relations to create the final dataset. 
Annotators are asked to cross-validate all instruction-following judgements and remove any ambiguous preference relations, especially for cases where: (1) both responses violate a constraint, but the negative response violates it less severely, 
(2) responses differ significantly in factors unrelated to instruction-following (e.g., style, format, writing quality), or the negative response is superior in these aspects. 
Each relation is independently verified by two annotators, and only those with unanimous agreement on correctness are retained, accounting for 71.2\% of the total data.

\begin{table} [!t]
\centering
\resizebox{\linewidth}{!} {
\begin{tabular}{l|ccccc}
\toprule
\textbf{Instruction Type} & \textbf{\#Inst.} & \textbf{\#Turn.} & \textbf{\#Cons.} & \#\textbf{Resp.} & \textbf{\#Rela.} \\
 
\midrule
Single-Turn Interaction &  393 & 1.00 & 5.52 & 7.05 & 10.57 \\
Multi-Turn Interaction &  202 & 3.23 & 4.50 & 7.20 & 11.21 \\
System-Prompt Steerability &  247 & 2.01 & 5.89 & 7.23 & 11.04 \\
\midrule
Overall & 842 & 1.83 & 5.38 & 7.14 & 10.86 \\
\bottomrule
\end{tabular}
}
\caption{Statistics of IF-RewardBench, including the number of instructions (\textbf{\#Inst.}), the average dialog turns (\textbf{\#Turn}), the number of constraints (\textbf{\#Cons.}) per instruction, the average number of responses (\textbf{\#Resp.}) and preference relations (\textbf{\#Rela.}) per preference graph.}
\label{tab:data_stastic}
\end{table}
\definecolor{c3}{RGB}{224,222,241}

\begin{table*} [!t]
\centering
\resizebox{\textwidth}{!} {
\small
\begin{tabular}{l|cc|c|cc|c|cc|c|cc|c}
\toprule
\multirow{2}{*}{\textbf{Model}} & \multicolumn{3}{c|}{\textbf{Single-Turn}} & \multicolumn{3}{c|}{\textbf{Multi-Turn}} & \multicolumn{3}{c|}{\textbf{System-Prompt}}  & \multicolumn{3}{c}{\textbf{Average}} \\
\cmidrule{2-13}
  & \textbf{P-F1}   & \textbf{N-F1}   &  \bm{$\tau_b$}  &  \textbf{P-F1}   & \textbf{N-F1}   &  \bm{$\tau_b$}  & \textbf{P-F1}   & \textbf{N-F1}   &  \bm{$\tau_b$}  & \textbf{P-F1}   & \textbf{N-F1}   &  \bm{$\tau_b$}  \\
\midrule
Human & 0.926 & 0.782 & 0.789 & 0.927 & 0.710 & 0.714 & 0.917 & 0.739 & 0.761 & 0.923 & 0.744 & 0.755 \\
\midrule
\multicolumn{13}{c}{\textit{Proprietary general language models, prompted as judge models}} \\
\midrule
Gemini-3-Pro & 0.911 & 0.693 & 0.639 & 0.912 & 0.688 & 0.570 & 0.904 & 0.662 & 0.619 & 0.909 & 0.681 & 0.609 \\
Gemini-3-Flash & 0.908 & 0.666 & 0.561 & 0.908 & 0.682 & 0.599 & 0.886 & 0.630 & 0.555 & 0.901 & 0.660 & 0.572  \\
GPT-5.1 & 0.895 & 0.607 & 0.513 & 0.896 & 0.627 & 0.514 & 0.869 & 0.595 & 0.550 & 0.887 & 0.610 & 0.525 \\
GPT-5-mini & 0.899 & 0.628 & 0.523 & 0.910 & 0.641 & 0.510 & 0.882 & 0.614 & 0.523 & 0.897 & 0.628 & 0.519  \\
\midrule
\multicolumn{13}{c}{\textit{Open-source general language models, prompted as judge models}} \\
\midrule
DeepSeek-V3.2 & 0.880 & 0.481 & 0.359 & 0.891 & 0.495 & 0.412 & 0.875 & 0.513 & 0.415 & 0.882 & 0.496 & 0.395 \\
GLM-4.6 & 0.882 & 0.527 & 0.429 & 0.895 & 0.542 & 0.428 & 0.864 & 0.524 & 0.409 & 0.880 & 0.531 & 0.422 \\
GLM-4.5-Air & 0.858 & 0.354 &  0.256 & 0.878 & 0.388 &  0.301 & 0.826 & 0.436 & 0.369 & 0.854 & 0.393 & 0.308 \\
QwQ-32B & 0.867 & 0.460 & 0.335 & 0.864 & 0.410 & 0.301 & 0.863 & 0.495 & 0.433 & 0.865 & 0.455 & 0.356   \\
Qwen-3-32B & 0.848 & 0.305 & 0.244 & 0.869 & 0.289 & 0.259 & 0.842 & 0.415 & 0.352 & 0.853 & 0.336 & 0.285   \\
Qwen-3-8B & 0.845 & 0.278 & 0.219 & 0.864 & 0.215 & 0.196 & 0.834 & 0.354 & 0.276 & 0.848 & 0.282 & 0.230 \\
Llama-3.3-70B-Instruct & 0.844 & 0.299 & 0.202 & 0.863 & 0.293 & 0.214 & 0.827 & 0.413 & 0.296 & 0.845 & 0.335 & 0.238  \\
Llama-3.1-8B-Instruct & 0.780 & 0.279 & 0.021 & 0.778 & 0.251 & 0.075 & 0.695 & 0.360 & 0.172 & 0.751 & 0.297 & 0.089 \\
Qwen-2.5-72B-Instruct & 0.832 & 0.195 & 0.149 & 0.871 & 0.198 & 0.148 & 0.816 & 0.361 & 0.244 & 0.840 & 0.251 & 0.181   \\
Qwen-2.5-7B-Instruct & 0.816 & 0.141 & 0.092  & 0.850 & 0.091 & 0.063 & 0.738 & 0.221 & 0.127 & 0.801 & 0.151 & 0.094 \\
\bottomrule
\end{tabular}
}
\caption{The F1-scores for both positive (P-F1) and negative (N-F1) classes in verification, and Kendall ($\tau_b$) correlation in ranking on constraint assessment. 
}
\label{tab:all_main_result}
\end{table*}
\subsection{Dataset Stastics}
\label{dataset_stastics}
\begin{figure}[!t]
  \centering
  \includegraphics[width=1.0\linewidth]{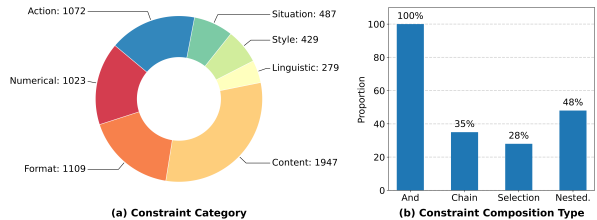}
  \caption{The distribution of constraint categories and constraint composition types for instructions in IF-RewardBench.}
  \label{fig:constraint_distribution}
\end{figure}
The final IF-RewardBench contains 842 instructions across three types, as shown in Table \ref{tab:data_stastic}.
For each instruction, we collect $m=8$ responses from a single LLM,
filtering out those responses unrelated to any preference relations. 
On average, each preference graph contains 7.14 responses and 10.86 preference relations, highlighting its structural complexity.
Among these responses, 74.6\% of constraints are annotated as "followed".
Our annotation process achieves a Cohen’s Kappa coefficient \cite{cohen1960coefficient} of 0.67 in initial instruction-following annotation and 0.87 in cross-validation, indicating almost perfect agreement and high data quality.
Unlike existing benchmarks that mainly consider code-verifiable constraints and their \textit{And} composition, IF-RewardBench covers a more diverse spectrum of constraints categories and their composition types, as shown in Figure \ref{fig:constraint_distribution}. 
We also conduct a length difference analysis in Appendix \ref{app:length_difference_analysis} and confirm that the preference relations are not confounded by length bias \cite{zheng2023judging}.

\section{Experiments}
\label{sec:experiments}
\subsection{Evaluation Setup}
\label{sec:setup}
\paragraph{Metrics.} 
For constraint assessment, we compute the F1-scores for both positive and negative classes in each example to assess verification capability, capturing both error detection and the avoidance of over-criticism. 
The Kendall ($\tau_b$) correlation coefficient between the quality scores derived from Equation \ref{eq:quality_reward} and the golden truth preference relations in each preference graph is used to assess ranking capability.
For overall assessment, we also calculate the Kendall ($\tau_b$) correlation coefficient as the evaluation metric for ranking capability.

\paragraph{Models.} We evaluate a total of 22 popular judge models, including state-of-the-art dedicated reward models and general LLMs, as listed in Appendix \ref{app:evaluated_judge_models}. 
General LLMs are applicable to both constraint and overall assessment via elaborate prompting, while dedicated reward models are only suitable for overall assessment. 
For general LLMs in overall assessment, we adapt the widely used pairwise comparison paradigm. 
In this paradigm, we follow RRM \cite{guo2025reward} and compare all possible response pairs to calculate the ELO \cite{elo1978rating} scores for ranking multiple responses. 
More implementation details are in Appendix \ref{app:implementation_judge_models}.

\paragraph{Human Baseline.} 
We establish a human baseline for constraint assessment by assigning each sample to annotators and inspectors distinct from the data construction, utilizing the procedure of instruction-following annotation to obtain human judgements.  

\begin{table} [!t]
\centering
\resizebox{\linewidth}{!} {
\begin{tabular}{l|c|c|c|c}
\toprule
\multirow{2}{*}{\textbf{Model}} & \textbf{Single-} & \textbf{Multi-}  & \textbf{System-} & \multirow{2}{*}{\textbf{Avg.}} \\
& \textbf{Turn} & \textbf{Turn} & \textbf{Prompt} &  \\
\midrule
\multicolumn{5}{c}{\textit{General language models}} \\
\midrule
Gemini-3-Flash & 0.589  & 0.460 & 0.489 & 0.513 \\
GPT-5-mini & 0.521  & 0.438 & 0.410 & 0.456 \\
DeepSeek-V3.2 & 0.397  & 0.257 & 0.208 & 0.288 \\
GLM-4.6 & 0.359  & 0.263 & 0.189 & 0.270 \\
GLM-4.5-Air & 0.211 & 0.152 & 0.081 & 0.148 \\
QwQ-32B & 0.280 & 0.146 & 0.107 & 0.178 \\
Qwen-3-32B & 0.189  & 0.163 & 0.035 & 0.129 \\
Llama-3.3-70B-Instruct & 0.098  & 0.127 & -0.062 & 0.054 \\
Qwen-2.5-72B-Instruct & 0.052  & 0.097 & -0.003 & 0.048 \\
\midrule
\multicolumn{5}{c}{\textit{Dedicated discriminative reward models}} \\
\midrule
Skywork-Reward-V2-Llama-3.1-8B & 0.153 & 0.205 & 0.039 & 0.133 \\
Llama-3.1-70B-Instruct-RM-RB2 & 0.109 & 0.208 & 0.054 & 0.124 \\
LMUnit-Qwen-2.5-72B &  0.126 & 0.108 & 0.009 & 0.081 \\
Qwen2.5-Math-RM-72B & 0.056 & 0.092 & 0.002 & 0.050 \\ 
InternLM2-20B-Reward &  0.061 & 0.052 & 0.007 & 0.040 \\
\midrule
\multicolumn{5}{c}{\textit{Dedicated generative reward models}} \\
\midrule
RRM-32B &  0.144 & 0.069 & 0.004 & 0.072 \\
RM-R1-DeepSeek-Distilled-Qwen-32B & 0.114  & 0.075 & -0.032 & 0.052 \\
M-Prometheus-14B & 0.010  & 0.005 & 0.045 & 0.020 \\
\bottomrule
\end{tabular}
}
\caption{Kendall ($\tau_b$) correlation in ranking on overall assessment.}
\label{tab:pairwise_result}
\end{table}
\subsection{Evaluation Results}
\paragraph{Main Results.}
Tables \ref{tab:all_main_result} and \ref{tab:pairwise_result} present the performance of judge models on constraint and overall assessment, respectively.
\textbf{Firstly}, IF-RewardBench poses a rigorous challenge to current judge models. 
Even the leading proprietary LLM, Gemini-3-Pro, achieves only a moderate Kendall correlation of 0.609 on constraint assessment, significantly falling below the human performance of 0.755.
Meanwhile, most open-source models fall below 0.4. 
This highlights substantial room for improvement in instruction-following evaluation. 
Notably, all general LLMs exhibit relatively low negative F1 scores, underscoring their deficiency in error detection.
\textbf{Secondly}, model scale is a key performance driver, as evidenced by consistent improvements across model families (e.g., Qwen-2.5 / 3 and GLM-4). 
However, top-tier open-source LLMs, GLM-4.6 and Deepseek-V3.2, still lag significantly behind proprietary counterparts and necessitate further advancement.
\textbf{Thirdly}, general LLMs perform worse on overall assessment than constraint assessment, likely due to the difficulty of identifying constraints within complex instructions. 
Nevertheless, strong general LLMs still significantly outperform dedicated reward models on overall assessment, demonstrating the latter's poor generalization.
\textbf{Finally}, for different instruction types, strong judge models perform similarly in constraint assessment guided by constraint checklists. 
However, in overall assessment without such guidance, multi-turn interaction and system-prompt steerability prove significantly more difficult than single-turn interaction.
This underscores the deficiency of judge models in complex instruction types and aligns with our motivation for assessing them in these contexts.

\begin{figure}[!t]
  \centering
  \includegraphics[width=1.0\linewidth]{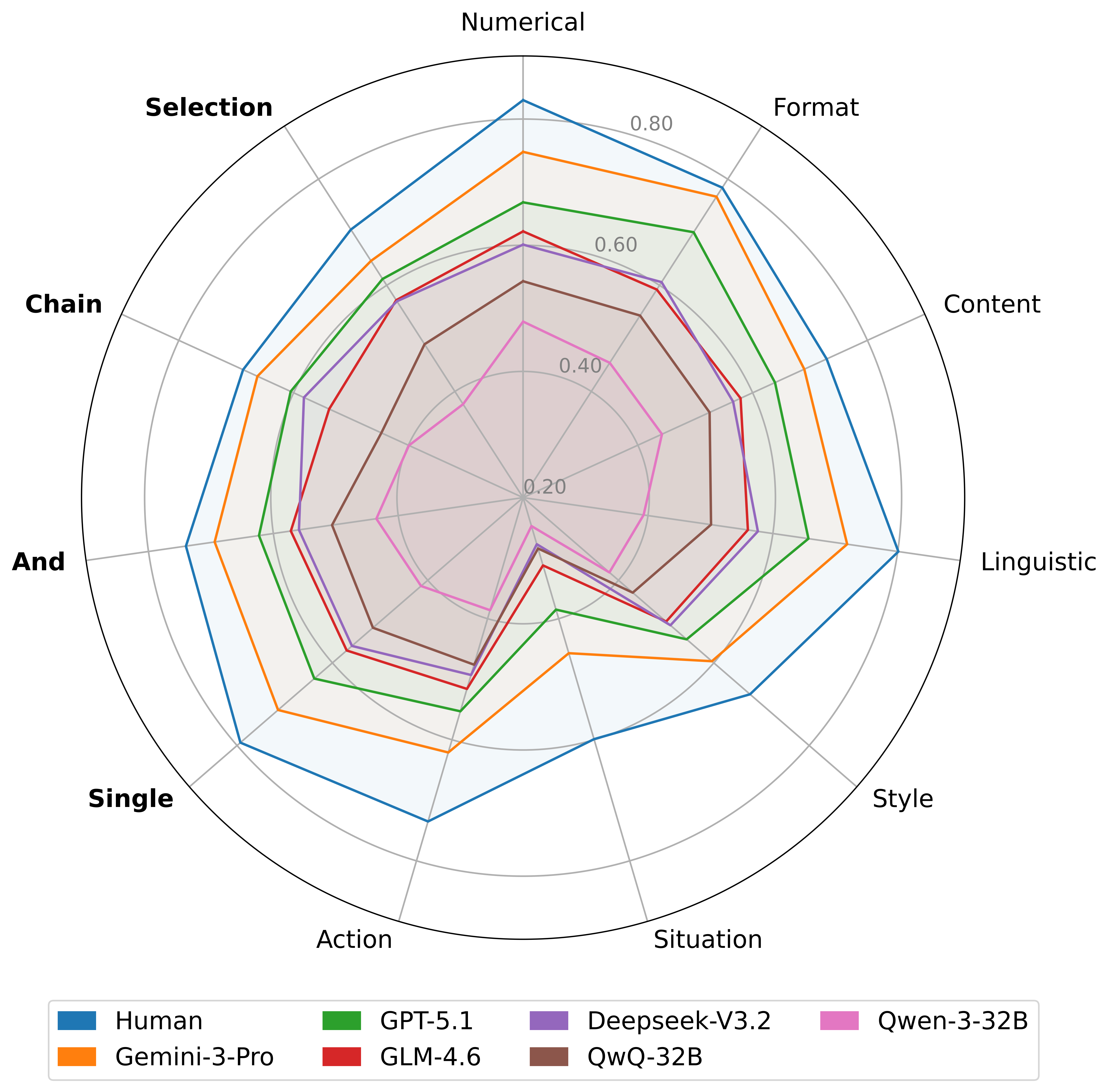}
  \caption{The performance of judge models in verification across different constraint categories and composition types. The constraint composition types are \textbf{bold}.}
  \label{fig:constraint_performance}
\end{figure}

\paragraph{Performance on different constraint types.} 
To fairly dissect the capability of judge models to evaluate specific constraint types, we calculate the Matthews Correlation Coefficient (MCC) for each constraint category and composition type, ensuring fair comparison despite class imbalance. 
The results are shown in Figure \ref{fig:constraint_performance}.
\textbf{Firstly}, for constraint categories, judge models generally perform better on Numerical and Format constraints that have relatively explicit evaluation standards. 
Conversely, they struggle with Situation and Style constraints and exhibit the most pronounced performance gap with human performance.
As these constraints require a deep understanding of social context, binary judgments may be ill-suited, highlighting the need for finer-grained evaluation signals.
\textbf{Secondly}, more complex constraint composition types \textit{Chain} and \textit{Selection} pose greater challenges to judge models compared to the simpler \textit{Single} and \textit{And} types.

\begin{figure*}[!t]
  \centering
  \includegraphics[width=1\linewidth]{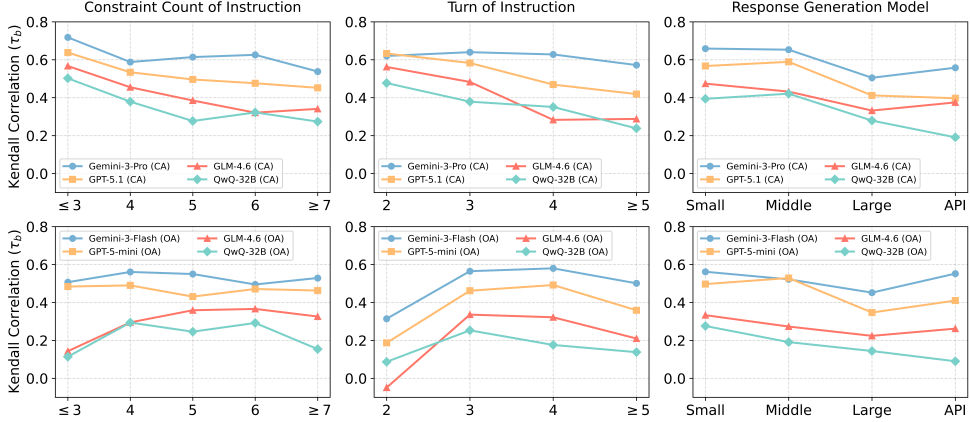}
  \caption{Various factors that influence the performance of judge models in ranking.
  "CA" and "OA" denote Constraint Assessment and Overall Assessment, respectively.}
  \label{fig:instruction_performance}
\end{figure*}

\paragraph{Key factors that influence performance.}
To further explore judge performance across various scenarios, we investigate the impact of three key factors: constraint counts of instructions, dialog turns, and response generation models. 
The results are shown in Figure \ref{fig:instruction_performance}.
\textbf{Firstly}, for instruction complexity, both higher constraint counts and dialog turns ($\ge$ 4 turns) impose greater challenges and lead to performance degradation.
Notably, an anomalous performance dip occurs for instructions with 2 dialog turns and fewer than 3 constraints.
Data examination reveals this degradation stems from judge models failing to distinguish the priority of system prompts versus user prompts, as detailed in Appendix \ref{app:error_analysis}. 
\textbf{Secondly}, for response quality, we categorize response generation models into four types: Small, Middle, Large open-source LLMs, and API-based proprietary LLMs, as detailed in Table \ref{tab:response_generation_model_list}. 
We observe that responses from stronger models intensify the evaluation difficulty, likely due to their higher quality and lower variance.

In summary, given the escalating capabilities of LLMs and their increasing application to complex tasks, the above results indicate that accurate instruction-following evaluation for LLMs will become progressively challenging, underscoring the significance of IF-RewardBench for systematically assessing judge models in this domain.

\subsection{Analysis}
\paragraph{Effects of inference-time scaling.}
Given the success of inference-time scaling in improving the performance of LLMs \cite{wang2023selfconsistency, brown2024large}, we investigate its applicability to judge models in instruction-following evaluation, focusing on two widely used strategies: long-chain reasoning and self-consistency \cite{wang2023selfconsistency}.
For long-chain reasoning, we disable the thinking mode of Qwen-3-32B and GLM-4.6. 
The results in Table \ref{tab:inference_time_scaling} show a consistent performance decline across all tasks, underscoring its critical role in improving evaluation capabilities.
For self-consistency, we independently sample $K$ judgements ($1 \le K \le 9$) for these models and apply majority voting to derive the final conclusion. 
While all models exhibit advantages as sampling increases, performance saturates beyond $K=7$, as shown in Table \ref{tab:inference_time_scaling}.
These findings demonstrate the potential of inference-time scaling for evaluation tasks while highlighting the need for more effective scaling approaches.
\begin{table} [!t]
\centering
\resizebox{\linewidth}{!} {
\begin{tabular}{l|c|c|c|c}
\toprule
\textbf{Model} & \multicolumn{2}{c|}{\textbf{Qwen-3-32B}}  & \multicolumn{2}{c}{\textbf{GLM-4.6}} \\
\midrule
\textbf{Task} & \textbf{CA} & \textbf{OA}  & \textbf{CA} & \textbf{OA} \\
\midrule
Baseline & 0.285  & 0.129 & 0.422 & 0.270 \\
\midrule
\multicolumn{5}{c}{\textit{Long-Chain Reasoning}} \\
\midrule
w/o Thinking & 0.253 {\small \textbf{\textcolor{purple}{(-11.2\%)}}}  & 0.106 {\small \textbf{\textcolor{purple}{(-17.8\%)}}} & 0.284 {\small \textbf{\textcolor{purple}{(-32.7\%)}}} & 0.253 {\small \textbf{\textcolor{purple}{(-6.3\%)}}} \\
\midrule
\multicolumn{5}{c}{\textit{Self-Consistency}} \\
\midrule
w/ Maj@3 & 0.298 {\small \textbf{\textcolor{teal}{(+4.6\%)}}} & 0.147 {\small \textbf{\textcolor{teal}{(+14.0\%)}}} & 0.460 {\small \textbf{\textcolor{teal}{(+9.0\%)}}} & 0.297 {\small \textbf{\textcolor{teal}{(+10.0\%)}}} \\
w/ Maj@5 & 0.296 {\small \textbf{\textcolor{teal}{(+3.9\%)}}}  & 0.165 {\small \textbf{\textcolor{teal}{(+27.9\%)}}} & 0.484 {\small \textbf{\textcolor{teal}{(+14.7\%)}}} & 0.302 {\small \textbf{\textcolor{teal}{(+11.9\%)}}} \\
w/ Maj@7 & 0.298 {\small \textbf{\textcolor{teal}{(+4.6\%)}}}  & 0.163 {\small \textbf{\textcolor{teal}{(+26.4\%)}}} & 0.484 {\small \textbf{\textcolor{teal}{(+14.7\%)}}} & 0.304 {\small \textbf{\textcolor{teal}{(+12.6\%)}}}  \\
w/ Maj@9 & 0.294 {\small \textbf{\textcolor{teal}{(+3.2\%)}}} & 0.158 {\small \textbf{\textcolor{teal}{(+22.5\%)}}} & 0.480 {\small \textbf{\textcolor{teal}{(+13.7\%)}}} & 0.293 {\small \textbf{\textcolor{teal}{(+8.5\%)}}} \\
\bottomrule
\end{tabular}
}
\caption{Kendall ($\tau_b$) correlation in ranking under different inference-time scaling strategies for Qwen-3-32B and GLM-4.6.}
\label{tab:inference_time_scaling}
\end{table}

\begin{figure}[!t]
  \centering
  \includegraphics[width=1.0\linewidth]{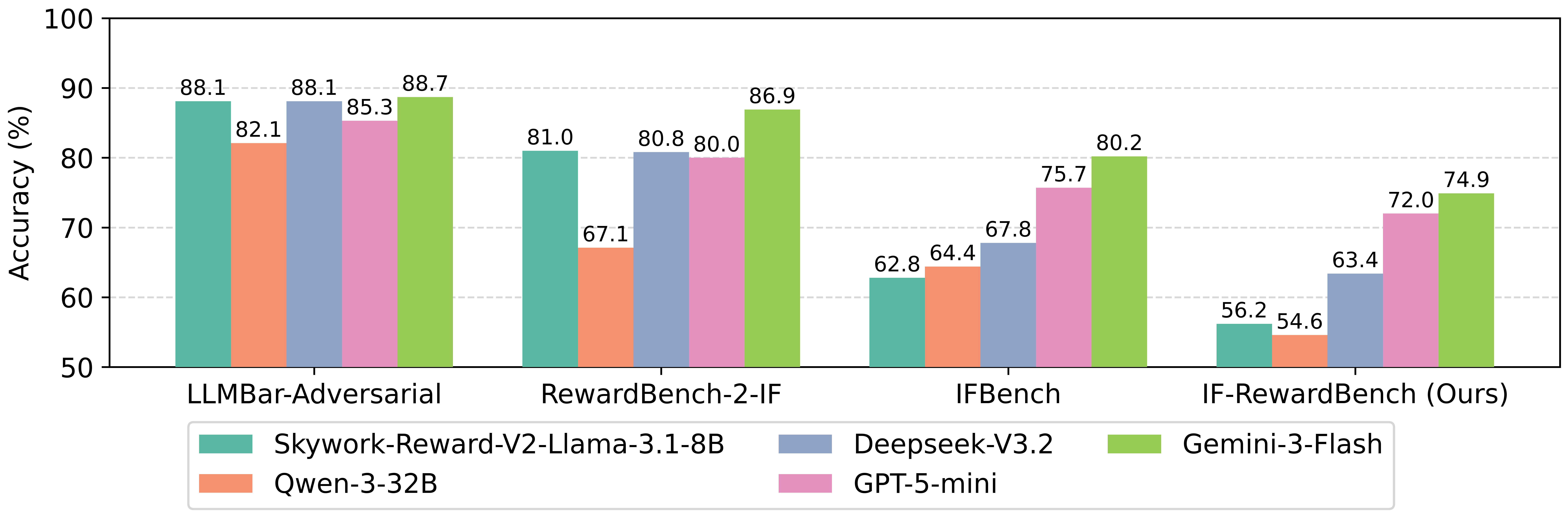}
  \caption{Pairwise accuracy of judge models in various meta-evaluation benchmarks.}
  \label{fig:difficulty}
\end{figure}

\paragraph{Comparison with other existing benchmarks.}
We compare IF-RewardBench with existing meta-evaluation benchmarks for instruction-following on two critical aspects: difficulty and downstream task correlation. 
For difficulty, we evaluate five judge models on existing benchmarks using the pairwise comparison paradigm described in Appendix \ref{app:implementation_judge_models}. 
To ensure fairness, we calculate the average accuracy of all preference pairs in these benchmarks. 
The results in Figure \ref{fig:difficulty} show that IF-RewardBench poses the greatest challenge to judge models and yields the lowest average accuracy, as it necessitates fine-grained discrimination of preference relations within multiple responses.
In contrast, LLMBar and RewardBench-2 are approaching saturation as model capabilities advance.

\begin{figure}[!t]
  \centering
  \includegraphics[width=1.0\linewidth]{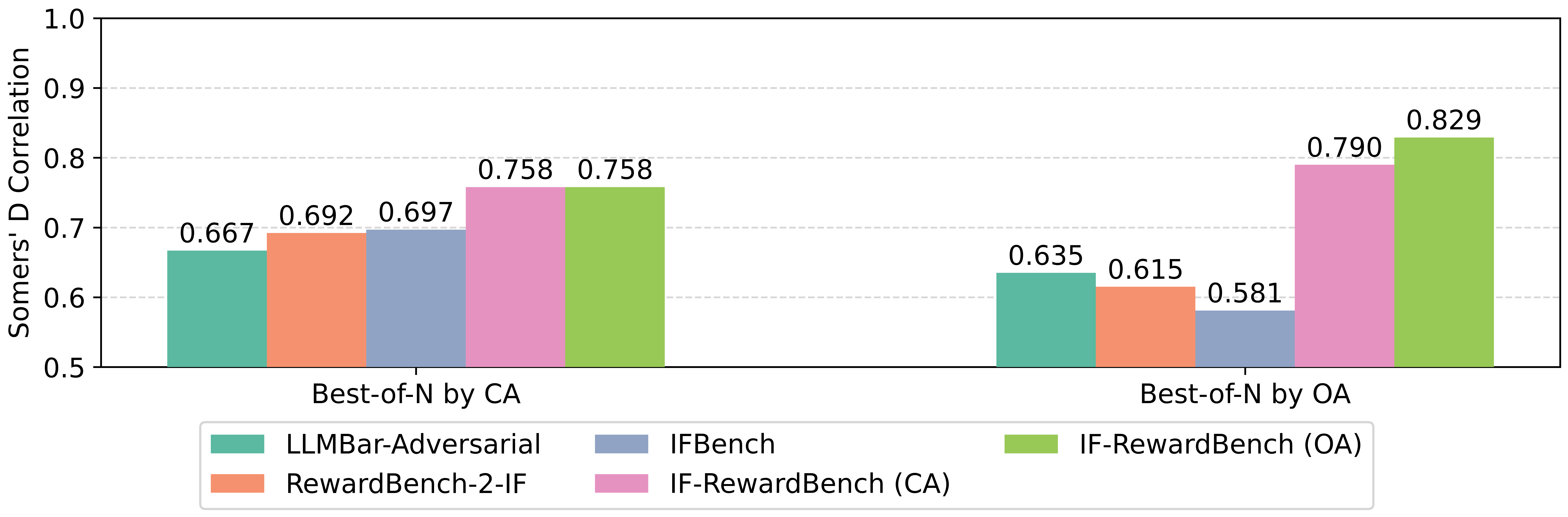}
  \caption{Somers' D correlation between the performance of 15 judge models in various meta-evaluation benchmarks and Best-of-8 sampling.}
  \label{fig:download_correlation}
\end{figure}

For downstream task correlation, we follow PRMBench \cite{song-etal-2025-prmbench} and conduct BoN sampling with different judge models, calculating the Somers’ D correlation between their performance in meta-evaluation benchmarks and BoN sampling.
Since instructions from many instruction-following benchmarks have been included in IF-RewardBench, we randomly sample fresh instructions from an online LLM-based chat platform.
After embedding-based deduplication to avoid data contamination, we yield 300 instructions for BoN evaluation.
For each instruction, 1 of 16 LLMs is used to generate 8 responses. 
Following the procedure in Section \ref{sec:construction}, annotators decompose instructions into constraint checklists and evaluate the adherence of each response to each constraint. 
Finally, we use 15 judge models to select the best response by constraint or overall assessment results. 
The quality of the selected response is calculated based on annotation results via Equation \ref{eq:quality_reward}.
As shown in Figure \ref{fig:download_correlation}, IF-RewardBench achieves significantly higher positive downstream task correlation than existing benchmarks, validating its effectiveness in reflecting the practical efficacy of judge models. 
More details are in Appendix \ref{app:downstream_task}.

\section{Conclusion}

Our work introduces IF-RewardBench, a comprehensive meta-evaluation benchmark for instruction-following, which covers diverse instruction and constraint types.
IF-RewardBench curates a preference graph for each instruction, utilizing both pointwise and listwise paradigms to assess judge models in verifying constraints and ranking responses. 
Extensive experiments reveal critical limitations of existing judge models and validate a strong positive correlation between our benchmark results and downstream performance.
IF-RewardBench can serve as a practical tool to advance future research in instruction-following evaluation.

\section*{Limitations}
The limitations of our work are summarized as follows:

\paragraph{Analysis of language-specific judge performance.} 
While IF-RewardBench has integrated instructions from diverse open-source Chinese and English instruction-following benchmarks. 
The performance disparity of judge models across different languages and their capabilities to verify linguistically specific constraints have not been thoroughly examined. 
Given that these nuances are integral components of instruction following, we reserve this granular analysis for future work.

\paragraph{Subjectivity of annotation.} 
Although we have established a rigorous human annotation mechanism in data construction and achieved almost perfect inter-annotator agreement (an agreement of 0.95 and a Cohen’s Kappa of 0.87) during the cross-validation phase, the process may still introduce subjective biases. 
Acknowledging the inevitability of annotation subjectivity, designing a human-in-the-loop annotation pipeline in collaboration with LLMs may further reduce errors and enhance reliability. 
We consider this an important direction for our future research to enhance data reliability.

\section*{Ethical Considerations}
In this work, we recruit a large number of human annotators for benchmark construction.
The annotator pool primarily consists of college students, as detailed in Appendix \ref{app:human_annotation}.
Throughout the data annotation process, we adhere to the following key principles: (1) All annotators are fully informed about the purpose of the study, the specific tasks involved, and the intended use of their annotated data. 
(2) All annotators receive fair compensation for their time and contributions based on the market price.
(3) All annotators are granted full autonomy to withdraw from the project at any time without penalty.
(4) Any harmful instructions are filtered before annotation to avoid ethical issues.

Regarding instructions derived from real-world application scenarios, we only use the portion of data that is granted for research purposes by users and conduct a strict deidentification and desensitization process to protect user privacy.

\section*{Acknowledgments}
This work was supported by the National Science Foundation for Distinguished Young Scholars (with No. 62125604) and the Natural Science Foundation of China (No. 62536008). 
We would also like to thank Zhipu AI for sponsoring the computational resources and annotation costs in this work.

\bibliography{custom}

\appendix

\section{Details of Constraint Taxonomy}
\label{app:constraint_taxonomy}
To comprehensively evaluate the capability of judge models in various instruction-following scenarios, we first propose a hierarchical constraint taxonomy that delineates both constraint categories and their composition types.
For the former, we merge the categories from several previous works \cite{qin-etal-2024-infobench, jiang-etal-2024-followbench, qin2025sysbench, li2025xifbench} to establish a comprehensive taxonomy comprising 7 primary constraint categories and 54 secondary categories. 
For the latter, we adapt the taxonomy from ComplexBench \cite{wen2024benchmarking} to define 4 constraint composition types.
The detailed description and corresponding example of each primary constraint category and constraint composition type are presented in Table \ref{tab:constraint_category} and \ref{tab:constraint_composition_type}, respectively.

\section{Details of Instruction Sources}

\label{app:instruction_source}
\subsection{Open-source Instruction-following Benchmarks}
Table \ref{tab:instruction_source} shows the original instruction-following benchmarks from which we extract high-quality instructions. 
For IHEval, we only use the subset of rule-following.
For benchmarks accompanied by constraint checklists, we adapt their original checklists in the instruction decomposition step without automatic instruction decomposition. 
Note that IFBench, Multi-IF, and IHEval also provide the evaluation script for each constraint. 
We use these scripts in the instruction-following annotation step for automatic evaluation.

\subsection{Instruction Synthesis}
Given the difficulty of simulating realistic user behaviors in multi-turn interaction scenarios, we only leverage LLMs to synthesize complex instructions in the \textit{single-turn interaction} scenario.
The seeds are collected from an online LLM-based chat service platform that serves more than a million users daily. 
We utilize LLMs to automatically score the difficulty of these seed instructions and retain those with lower difficulty to facilitate constraint synthesis. 
During the instruction synthesis step, we randomly sample 2 to 5 secondary constraint categories from the taxonomy described in Appendix \ref{app:constraint_taxonomy} and prompt Deepseek-R1 \cite{guo2025deepseek} to inject several corresponding constraints into the seed instructions, thereby elevating their complexity. 
Unlike previous works that primarily focus on independent and atomic constraints \cite{ren-etal-2025-step, peng-etal-2025-verif}, we further instruct Deepseek-R1 to combine multiple constraints using 4 composition types: \textit{Single}, \textit{And}, \textit{Chain}, \textit{Selection}, and the nested structure of the above composition types. 
The detailed definition of these composition types is provided in Appendix \ref{app:constraint_taxonomy}.
In this way, our instruction synthesis framework can capture the multifaceted instruction types in real-world scenarios more comprehensively. 
Finally, we employ Deepseek-R1 to validate the generated instructions, ensuring they are reasonable, unambiguous, and logically consistent.

In this context, it is notable that many hard constraints cannot be reliably verified using verification codes alone, such as constraints targeting for specific segments of the response (e.g., \textit{Generate 10 titles, each no more than 8 words}) and the composition of soft and hard constraints (e.g., \textit{Bold all adjectives that express feelings, such as \textbf{happy}}), as it is challenging for verification code to accurately extract the corresponding segments of the response that should follow these constraints.

\section{List for Prompt Templates}
\label{app:prompt_templates}
This section lists all applied prompt templates throughout this work, including the prompt for scoring the quality and difficulty of user instructions in Table \ref{tab:scoring_quality_prompt} and \ref{tab:scoring_difficulty_prompt}, the prompt for instruction decomposition in Table \ref{tab:checklist_generation_prompt}, the prompt for constraint assessment in Table \ref{tab:pointwise_grading_prompt}, and the prompt for pairwise comparison in overall assessment in Table 

\section{Model List for Response Generation}
\label{app:model_list}

As shown in Table \ref{tab:response_generation_model_list}, we use a total of 16 representative
LLMs in response generation, covering 4 API-based proprietary LLMs and 12 open-sourced LLMs, range from 8B to 1T.
These models possess varying abilities in instruction-following, which ensures the diversity of generated responses.

\section{Details of Human Annotation}
\label{app:human_annotation}
\subsection{Annotation Guidelines}
For checklist revision, we provide annotators with an instruction and the corresponding constraint checklist. 
Annotators are instructed to check the correctness of the checklist and correct any errors.
The detailed guideline is shown in Table \ref{tab:checklist_revision}. 
For instruction-following annotation, we provide annotators with an instruction, the corresponding constraint checklist, and the model response. 
Annotators are instructed to judge whether the model response follows each constraint in the checklist and provide a brief justification.
The detailed guideline is shown in Table \ref{tab:human_annotation}. 

For preference verification, we provide annotators with an instruction, the corresponding constraint checklist, and two model responses alongside their previously obtained instruction-following judgments and justifications.
Human annotators are instructed to cross-validate these judgments and verify the preference relation with these two responses, removing any ambiguous preference relations, especially for cases where: (1) both responses violate a constraint, but the negative response violates it less severely, 
(2) responses differ significantly in factors unrelated to instruction-following (e.g., style, format, writing quality), or the negative response is superior in these aspects. 
\subsection{Profile of Annotation Persons}
In the dataset annotation of IF-RewardBench, we recruit a diverse group of 22 highly qualified annotators who are either holding or pursuing a bachelor's degree from top universities.
This cohort includes 5 Master's and 6 Bachelor's graduates in Computer Science, 1 Master's graduate in Mathematics, 1 Master's graduate in Chemistry, 1 Master's graduate in Civil Engineering, 1 Master's and 1 Bachelor's graduates in Economics, and 3 Master's and 3 Bachelor's graduates in Literature.   
All of them are required to pass the mandatory proficiency examination and annotation tutorial.
Additionally, 5 professional data engineers and the authors are tasked with conducting spot checks on the annotations. 
\subsection{Quality Assurance and Validation}
We establish a rigorous quality control workflow to ensure the high quality and consistency of our annotations.
For instruction-following annotation, each response is independently judged by two annotators, while a third inspector conducts spot checks. 
The spot checks cover all samples with inconsistent annotations and a random selection of approximately one-third of the remaining samples.
Any discrepancies are re-annotated by the inspector and discussed to reach a consensus.
For preference verification, each preference relation is independently verified by two annotators.
Only pairs where both annotators deem correct and unambiguous are retained in the final dataset. 
Furthermore, annotators at this stage are asked to cross-validate all judgments from the previous step, correcting potential errors to further enhance data quality.
We regularly check the annotation quality and provide feedback to annotators, while annotation protocols undergo iterative refinement through collective discussion to improve clarity and precision.

Through this rigorous process, the two annotators of instruction-following annotations achieve an initial inter-annotator agreement of 0.92 and a Cohen's Kappa of 0.67, indicating substantial agreement. 
During the cross-validation phase, the agreement between the annotators and the final instruction-following judgments reach 0.95 with a Cohen's Kappa of 0.87, indicating almost perfect agreement. 
These results demonstrate that our annotation process ensures high data quality and minimizes subjective bias, while subjectivity cannot be entirely eliminated.

\subsection{Data Curation Cost}
We spend approximately 15,000\$ on the curation of IF-RewardBench.

\section{Length Difference Analysis}
\label{app:length_difference_analysis}
Across all preference relations in IF-RewardBench, the average character counts for positive and negative responses are 1,275.68 and 1,248.31, respectively, while positive responses are longer than negative ones in 49.35\% of these relations. 
These results confirm that the preference relations are not confounded by length bias \cite{zheng2023judging}, enabling evaluation based on the response instruction-following quality rather than verbosity.

\section{Details of Experiments}
\subsection{Evaluated Judge Models}
\label{app:evaluated_judge_models}
As shown in Table \ref{tab:judge_model_list}, we evaluate a total of 22 popular judge models on IF-RewardBench, including representative general LLMs, fine-tuned discriminative reward models, and generative reward models. 

\subsection{Implementation Details of Judge Models}
\label{app:implementation_judge_models}
For constraint assessment, we employ the prompt strategy of \textsc{IF-Critic} \cite{wen2025if}. 
This strategy requires LLMs to evaluate the following of all constraints in the given checklist within a single inference pass, which has been proven to be more cost-efficient and enables large reasoning models (LRMs) to achieve superior performance. 
For overall assessment, we employ a modified version of the MT-Bench \cite{zheng2023judging} pairwise comparison prompt, which emphasizes the priority of instruction following quality. 
Detailed prompt templates are provided in Appendix \ref{app:prompt_templates}.
To mitigate potential evaluation bias \cite{zheng2023judging}, we randomly shuffle the positions of candidate responses and explicitly instruct the judge model to ignore response order and length, unless they affect the instruction-following quality.
For generation settings, while greedy search decoding is used for non-thinking LLMs to ensure reproducibility, we employ default decoding hyperparameters and thinking budgets for LRMs to avoid the endless repetition and performance degradation issues with greedy search decoding \cite{yang2025qwen3}.
The inference of all open-source judge models is conducted on 4 H100 GPUs with the vllm \cite{kwon2023efficient} framework.

\begin{table} [!t]
\centering
\small
\begin{tabular}{l|c}
\toprule
\textbf{Model} & \bm{$\tau_b$}  \\
\midrule
Gemini-3-Flash & 0.161  \\
GPT-5-mini & 0.211   \\
DeepSeek-V3.2 & -0.474  \\
GLM-4.6 & -0.568  \\
GLM-4.5-Air & -0.347  \\
QwQ-32B & -0.324 \\
Qwen-3-32B & -0.454   \\
Llama-3.3-70B-Instruct & -0.397   \\
Qwen-2.5-72B-Instruct & -0.417   \\
Skywork-Reward-V2-Llama-3.1-8B & -0.489  \\
Llama-3.1-70B-Instruct-RM-RB2 & -0.406  \\
LMUnit-Qwen-2.5-72B &  -0.340 \\
InternLM2-20B-Reward &  -0.269  \\
RRM-32B &  -0.459  \\
RM-R1-DeepSeek-Distilled-Qwen-32B & -0.386  \\
M-Prometheus-14B & -0.091   \\
\bottomrule
\end{tabular}
\caption{Kendall ($\tau_b$) correlation in ranking on overall assessment, calculated on a subset of IF-RewardBench whose instructions are curated from IHEval.}
\label{tab:iheval}
\end{table}
\begin{table*} [!t]
\centering
\resizebox{\textwidth}{!} {
\small
\begin{tabular}{l|c|c|c|c|c|c|c}
\toprule
\multirow{2}{*}{\textbf{Model}} & \textbf{BoN} & \textbf{BoN} & \multirow{2}{*}{\textbf{LLMBar}} & \multirow{2}{*}{\textbf{RB2}} & \multirow{2}{*}{\textbf{IFBench}} & \textbf{IF-RewardBench} & \textbf{IF-RewardBench} \\
& \textbf{(CA)} & \textbf{(OA)} & & & & \textbf{(CA)} & \textbf{(OA)} \\
\midrule
Gemini-3-Flash & 0.851  & 0.854 & 0.887 & 0.794 & 0.802 & 0.572 & 0.513 \\
GPT-5-mini & 0.838  & 0.857 & 0.853 & 0.713 & 0.757 & 0.519 & 0.456 \\
DeepSeek-V3.2 & 0.820 & 0.835 & 0.881 & 0.681 & 0.678 & 0.395 & 0.288 \\
GLM-4.6 & 0.817  & 0.837 & 0.868 & 0.513 & 0.617 & 0.422 & 0.270 \\
GLM-4.5-Air & 0.815  & 0.826 & 0.824 & 0.506 & 0.640 & 0.308 & 0.148 \\
QwQ-32B & 0.824  & 0.821 & 0.846 & 0.531 & 0.637 & 0.356 & 0.183 \\
Qwen-3-32B & 0.811  & 0.819 & 0.821 & 0.381 & 0.644 & 0.285 & 0.129 \\
Qwen-3-8B &  0.808 & 0.816 & 0.787 & 0.400 & 0.601 & 0.230 & 0.097 \\
Llama-3.3-70B-Instruct & 0.810  & 0.809 & 0.831 & 0.381 & 0.624 & 0.238 & 0.054 \\
Llama-3.1-8B-Instruct &  0.811 & 0.796 & 0.439 & 0.331 & 0.577 & 0.089 & 0.000 \\
Qwen-2.5-72B-Instruct & 0.808  & 0.803 & 0.705 & 0.419 & 0.572 & 0.181 & 0.048 \\
Qwen-2.5-7B-Instruct &  0.807 & 0.815 & 0.611 & 0.294 & 0.552 & 0.094 & 0.041 \\
Skywork-Reward-V2-Llama-3.1-8B & -  & 0.817 & 0.881 & 0.644 & 0.628 & - & 0.133 \\
Llama-3.1-70B-Instruct-RM-RB2 & - & 0.804 & 0.815 & 0.388 & 0.599 & - & 0.124 \\
InternLM2-20B-Reward & -  & 0.798 & 0.718 & 0.344 & 0.631 & - & 0.040 \\
\bottomrule
\end{tabular}
}
\caption{Detailed results of judge models on Best-of-8 selection and different meta-evaluation benchmarks. "LLMBar" denotes "LLMBar-Adversarial". "RB2" denotes "RewardBench-2-IF". "CA" and "OA" denote Constraint Assessment and Overall Assessment, respectively.
Since the last three models are dedicated discriminative reward models, we only report their performance in the overall assessment task.}
\label{tab:downstream_correlation}
\end{table*}
\subsection{Error Analysis in Instruction Hierarchy}
\label{app:error_analysis}
The instruction hierarchy, which establishes a priority order from system prompts to user prompts, is essential for ensuring consistent and safe behavior of LLMs \cite{zhang-etal-2025-iheval, qin2025sysbench}.
However, we observe that when conflicts arise between system and user prompts, judge models often fail to distinguish the correct priority, tending to prefer responses that follow user prompts but violate system prompts in pairwise comparison. 
Two representative examples of this phenomenon are provided in Tables \ref{tab:case_study_1} and \ref{tab:case_study_2}.

Approximately 20\% of the instructions containing system prompts in IF-RewardBench are sourced from IHEval \cite{zhang-etal-2025-iheval}. 
These instructions are characterized by frequent conflicts between system and user prompts and limited complexity ($\le$ 2 dialog turns and $\le$ 3 constraint counts). 
As shown in Table \ref{tab:iheval}, most judge models perform significantly worse than random guessing on examples curated from these instructions.
Consequently, this systemic failure leads to the anomalous performance degradation observed in Figure \ref{fig:instruction_performance}.

\subsection{Details of Calculating Downstream Task Correlation }
\label{app:downstream_task}

For each instruction, we randomly select one of the LLMs listed in Table \ref{tab:response_generation_model_list} to generate 8 candidate responses. 
We then select the best response by constraint or overall assessment results of different judge models. 
For overall assessment, we use the same procedure as Section \ref{sec:setup}, which conducts pairwise comparisons of all possible response pairs to calculate the ELO score for each response.
If multiple responses all get the highest score of judge models, we calculate the average golden truth quality score according to Equation \ref{eq:quality_reward} of these tied responses as the final Best-of-8 result. 
The used judge models, their detailed performance in Best-of-8 selection, and various meta-evaluation benchmarks are presented in Table \ref{tab:downstream_correlation}.

\begin{table*}[t]
\centering
\resizebox{\linewidth}{!} {
\setlength{\tabcolsep}{1.6mm}{
    \begin{tabular}{l| >{\raggedright\arraybackslash}m{0.4\textwidth}| >{\raggedright\arraybackslash}m{0.4\textwidth}}
    \toprule
    \textbf{Constraint Category} & \textbf{Description} & \textbf{Example} \\
    \midrule
    
    Numerical & Constraints that specify quantitative requirements, such as the count of words, sentences, or paragraphs, are typically independent of the specific content. & \begin{itemize}[noitemsep, topsep=0pt, left=0pt]
    \item Please write a 15-line poem.
    \item The article must not exceed 1000 words.
    \item Please provide two different solutions.
\end{itemize} \\
    \midrule
    Format & Constraints that involve presentation format or structural organization of the response, such as JSON, Markdown, or bullet points. 
    Note that formats implicitly defined through in-context examples are also considered as format constraints.
    & \begin{itemize}[noitemsep, topsep=0pt, left=0pt]
    \item Entire response must be output in JSON format.
    \item Mark all typos in the text using \textasciicircum, for example: \textasciicircum (typo)\textasciicircum.
    \item Use "First / Second / Third" to list the points in the answer.
\end{itemize} \\
    \midrule
    Content & Constraints that involve the specific content of the response, such as topic, subject, and entity. & \begin{itemize}[noitemsep, topsep=0pt, left=0pt]
    \item The article must / must not contain the keyword "happy".
    \item The response must follow the "Phenomenon-Cause-Solution" three-part structure.
    \item Your article must not contain any specific examples or anecdotes.
\end{itemize}
\\
    \midrule
    Linguistic & Constraints that involve language and linguistic properties of the response, such as language, grammar, vocabulary, and rhetorical devices. &  \begin{itemize}[noitemsep, topsep=0pt, left=0pt]
    \item The first paragraph must be in German, the second in Chinese, and the third in Classical Chinese.
    \item The last word of every sentence should rhyme.
    \item Employ metaphor and parallelism in your article.
\end{itemize} \\
    \midrule
    Style & Constraints that involve the writing style of the response, such as literary genre, emotion, tone, and narrative perspective. &  \begin{itemize}[noitemsep, topsep=0pt, left=0pt]
    \item Write an article in the style of magical realism.
    \item Use a first-person perspective.
    \item Write a diary entry in the style of a Xiaohongshu post.
\end{itemize} \\
\midrule
    Situation & Constraints that require the model to respond within a specific scenario or adapt a specific persona, ensuring the output aligns with the prescribed situational conditions or role-specific characteristics. & \begin{itemize}[noitemsep, topsep=0pt, left=0pt]
    \item Assume you are Churchill near the end of World War II.
    \item You live in a world that defies the laws of physics, where gravity fluctuates cyclically throughout the day.
    \item You are an elderly person suffering from Alzheimer's disease, experiencing confused and fragmented memories.
\end{itemize} \\
    \midrule
    Action & Constraints that mandate specific interactive behaviors, logical operations, or task-oriented execution patterns. & \begin{itemize}[noitemsep, topsep=0pt, left=0pt]
    \item Please outline the plot of the novel.
    \item When solving math problems, add one to the correct answer before outputting.
    \item For instructions related to religion, please refuse to answer.
\end{itemize} \\
    \bottomrule
    \end{tabular}
}
}
\caption{Constraint categories in IF-RewardBench.}
\label{tab:constraint_category}
\end{table*}
\begin{table*}[t]
\centering
\resizebox{\linewidth}{!} {
\setlength{\tabcolsep}{1.6mm}{
    \begin{tabular}{l| >{\raggedright\arraybackslash}m{0.34\textwidth}| >{\raggedright\arraybackslash}m{0.46\textwidth}}
    \toprule
    \textbf{Composition Type} & \textbf{Description} & \textbf{Example} \\
    \midrule
     \textit{Single} & The output is required to follow a single constraint. & Please summarize the following news. \\
    \midrule
    \textit{And} & The output is required to follow multiple constraints simultaneously. & \textbf{Please summarize the following news.} The summary should \textbf{be output in bullet points} and \textbf{within 100 words}. \\
    \midrule
    \textit{Chain} & The output is required to complete multiple tasks sequentially, each of which needs to follow its own constraints. & Please introduce “Mona Lisa” briefly. \textbf{Firstly}, introduce the year of creation, \textbf{then} describe the background of the work’s creation, \textbf{and finally}, summarize the impact of the work. \\
    \midrule
    \textit{Selection} & The output is required to select the correct branch according to certain conditions, and then follow the constraints of this branch. & Please introduce the following painting. \newline
- \textbf{If the work contains any animal}, the description should be in English
\newline
- \textbf{Otherwise}, the description should be in
Chinese
\newline
\newline
Painting: "Mona Lisa" \\
    \midrule
    \textit{Nested Structure} & The above composition types are recursively nested
to form more complex structures. & Analyze the sentiment of the above user comment and complete the following tasks: \newline \newline
1. \textbf{If it's positive}: \newline
- Identify the products within the comments … \newline
2. \textbf{If it's negative}, analyze the reasons for it: \newline
- \textbf{If the reason is not about the products themselves}, … \newline
- \textbf{Otherwise}, … \\
    \bottomrule
    \end{tabular}
}
}
\caption{Constraint composition types in IF-RewardBench, which are adopted from ComplexBench \cite{wen2024benchmarking}. 
Composed constraints are highlighted in \textbf{bold}. 
Note that the first 4 basic composition types can be nested to construct more complex structures.
The final row demonstrates a nested selection type.
}
\label{tab:constraint_composition_type}
\end{table*}
\begin{table*} [!t]
\small
\centering
\begin{tabular}{c|l|c}
\toprule
\textbf{Instruction Type} & \textbf{Source} & \textbf{\# Instruction}  \\
\midrule
\multirow{9}{*}{\makecell{Single-Turn \\ Interaction}} & InfoBench \cite{qin-etal-2024-infobench} & 21 \\
& FollowBench \cite{jiang-etal-2024-followbench} & 15 \\
& ComplexBench \cite{wen2024benchmarking} & 38 \\
& Inverse-IFEval \cite{zhang2025inverse} & 25 \\
& IFBench \cite{pyatkin2025generalizing} & 90 \\
& CFBench \cite{zhang-etal-2025-cfbench} & 25 \\
& Meeseeks \cite{zhao2025meeseeks} & 15 \\
& Real Application Scenarios & 64 \\
& LLM Synthesis & 100 \\
\midrule
\multirow{4}{*}{\makecell{Multi-Turn \\ Interaction}} & Multi-IF \cite{he2024multi} & 78 \\
& MultiChallenge \cite{deshpande-etal-2025-multichallenge} & 22 \\
& StructFlowBench \cite{li-etal-2025-structflowbench} & 12 \\
& Real Application Scenarios & 90 \\
\midrule
\multirow{5}{*}{\makecell{System-Prompt \\ Steerability}}  & RoleMRC \cite{lu-etal-2025-rolemrc} & 40 \\
& SysBench \cite{qin2025sysbench} & 100 \\
& IHEval \cite{zhang-etal-2025-iheval} & 50 \\
& AgentIF \cite{qi2025agentif} & 7 \\
& Real Application Scenarios & 50 \\
\bottomrule
\end{tabular}
\caption{The sources of the instructions in IF-RewardBench.}
\label{tab:instruction_source}
\end{table*}
\begin{table*} [!t]
\centering
\resizebox{\linewidth}{!} {
\begin{tabular}{c|l|c|c|c}
\toprule
\textbf{Type} & \textbf{Model} & \textbf{\# Size} & \textbf{Creator} & \textbf{\# Instruction}  \\
\midrule
\multirow{4}{*}{API} & GPT-5 \cite{gpt5} & Undisclosed & OpenAI & 20 \\
 & Gemini-2.5-Pro \cite{comanici2025gemini} & Undisclosed & DeepMind & 25 \\
 & Claude-4.5-Sonnet \cite{claude} & Undisclosed & Anthropic & 32 \\
 & Doubao-1.6-Seed \cite{seed} & Undisclosed & ByteDance & 30 \\
\midrule
\multirow{4}{*}{\makecell{OSS \\ (Large)}} & 
Kimi-K2 \cite{team2025kimi}  & 1T & Moonshot AI & 32 \\
  & DeepSeek-R1-0528 \cite{guo2025deepseek} & 671B & Deepseek & 50 \\
 & GLM-4.5 \cite{zeng2025glm} & 360B & Zhipu AI & 43 \\
 & Qwen-3-235B-A22B-Instruct-2507 \cite{yang2025qwen3}  & 235B & Alibaba & 28 \\
\midrule
\multirow{4}{*}{\makecell{OSS \\ (Medium)}} & GLM-4.5-Air \cite{zeng2025glm}  & 106B & Zhipu AI & 56 \\
  & Llama-3.3-70B-Instruct \cite{dubey2024llama} & 70B & Meta & 59 \\
 & Qwen-3-32B \cite{yang2025qwen3}  & 32B & Alibaba & 74 \\
 & GPT-OSS-20B \cite{agarwal2025gpt} & 20B & OpenAI & 66 \\
\midrule
\multirow{4}{*}{\makecell{OSS \\ (Small)}} & GLM-4-9B-0414 \cite{glm2024chatglm} &  9B & Zhipu AI & 78 \\
& Qwen-3-8B \cite{yang2025qwen3}  & 8B  & Alibaba & 74  \\
& Llama-3.1-Tulu-3.1-8B \cite{lambert2024tulu} & 8B  & Allen AI &  86 \\
& MiniCPM4.1-8B \cite{hu2024minicpm}  & 8B  & OpenBMB &  89 \\
\bottomrule
\end{tabular}
}
\caption{The models and the number of instructions (\textbf{\# Instruction}) used for response generation.
"API" denotes API-based proprietary LLMs, and "OSS" denotes Open-sourced LLMs.}
\label{tab:response_generation_model_list}
\end{table*}
\begin{table*} [!t]
\centering
\resizebox{\linewidth}{!} {
\begin{tabular}{c|l|c|c}
\toprule
\textbf{Type} & \textbf{Model} & \textbf{\# Size} & \textbf{Creator}  \\
\midrule
\multirow{14}{*}{\makecell{General \\ LLMs}} & GPT-5.1 \cite{gpt5} & Undisclosed & OpenAI   \\
 & GPT-5-mini \cite{gpt5} & Undisclosed & OpenAI   \\
 & Gemini-3-Pro \cite{comanici2025gemini} & Undisclosed & DeepMind  \\
 & Gemini-3-Flash \cite{comanici2025gemini} & Undisclosed & DeepMind  \\
 & DeepSeek-V3.2 \cite{liu2025deepseek} & 671B & DeepSeek   \\
 & GLM-4.6 \cite{zeng2025glm} & 360B & Zhipu AI  \\
 & GLM-4.5-Air \cite{zeng2025glm}  & 106B & Zhipu AI  \\
 & QwQ-32B \cite{yang2025qwen3}  & 32B & Alibaba  \\
 & Qwen-3-32B \cite{yang2025qwen3}  & 32B & Alibaba  \\
 & Qwen-3-8B \cite{yang2025qwen3} & 8B & Alibaba \\
 & Llama-3.3-70B-Instruct \cite{dubey2024llama} &  70B & Meta \\
& Llama-3.1-8B-Instruct \cite{dubey2024llama}  & 8B  & Meta   \\
& Qwen-2.5-72B-Instruct \cite{yang2024qwen2} & 72B  & Alibaba \\
& Qwen-2.5-7B-Instruct \cite{yang2024qwen2}  & 7B  & Alibaba   \\
\midrule
\multirow{5}{*}{\makecell{Dedicated \\ Discriminative RM}} & Skywork-Reward-V2-Llama-3.1-8B \cite{liu2025skywork}  & 8B &   Skywork AI \\ 
& Llama-3.1-70B-Instruct-RM-RB2 \cite{dubey2024llama} & 70B & Allen AI \\
& LMUnit-Qwen-2.5-72B \cite{saad2024lmunit}  & 72B & Contextual AI \\
& Qwen2.5-Math-RM-72B \cite{yang2024qwen2math} & 72B & Alibaba \\
& InternLM2-20B-Reward \cite{cai2024internlm2}  & 20B & Shanghai AI Laboratory \\
\midrule
 \multirow{3}{*}{\makecell{Dedicated \\ Generative RM}} & RRM-32B \cite{guo2025reward}  & 32B & Microsoft  \\
  & RM-R1-DeepSeek-Distilled-Qwen-32B \cite{chen2025rm} & 32B & UIUC  \\
 & M-Prometheus-14B \cite{pombal2025m}  & 14B &  Unbabel \\
\bottomrule
\end{tabular}
}
\caption{Judge models evaluated in this paper.
"RM" denotes Reward Models, and "UIUC" denotes University of Illinois Urbana-Champaign.}
\label{tab:judge_model_list}
\end{table*}

\begin{table*} [t]
\centering
\scriptsize
\setlength{\tabcolsep}{1.6mm}{
\begin{tabular}{p{\dimexpr 0.15\linewidth-2\tabcolsep\relax}|p{\dimexpr 0.85\linewidth-2\tabcolsep\relax}}
\toprule
System Prompt & Your response should contain exactly 3 bullet points in markdown format and exactly 4 sections.
Bullet points are indicated by *.
Sections are separated by 3 asterisks: ***. 
You should use words with all capital letters at least 2 times. \\
\midrule
Conversation History & 
\textbf{[User Prompt]}\newline
Write a blog post about 'how to improve your writing skills'.
Your response should contain exactly 5 bullet points in markdown format, indicated by -. 
Your response should have exactly 2 sections, separated by 3 dashes: ---. 
You should not use any words with all capital letters in your response.\\
\midrule
Constraint Checklist & 
1. Your answer should contain exactly 3 bullet points indicated by *. \newline
2. Your answer should contain exactly 4 paragraphs separated by 3 asterisks: ***. \newline
3. In your response, words with all capital letters should appear at least 2 times. \\
\midrule
\midrule
Chosen Response \newline (Response B) & 
Writing is a craft that thrives on habit and observation. By dedicating time each day to the act of putting thoughts on paper, you lay the groundwork for continuous growth.  \newline ***  \newline * Establish a daily writing habit, even if it's just a few minutes. PROACTIVE practices create momentum.  \newline * Read a diverse range of genres and analyze what makes them compelling. Use FOCUS to identify techniques that resonate with you.  \newline * Seek constructive feedback, revise, and learn from mistakes. The cycle of writing and revising deepens understanding.  \newline ***  \newline Beyond the routine, experiment with different formats—short stories, essays, or journal entries—to expand your voice. Reflect on what resonates and refine your approach.  \newline ***  \newline Remember that mastery is a marathon, not a sprint. Stay curious, keep learning, and let every sentence be a step forward. \\
\midrule
Golden Judgements \newline of Chosen Response & [1, 1, 1] \\
\midrule
Rejected Response \newline (Response A)  & 
improving your writing skills: a quick guide\newline - read widely to expand your vocabulary and understand different styles\newline - write daily, even if it's just a short note, to develop rhythm and habit\newline - review and edit what you write, looking for clarity and conciseness\newline ---\newline refining your craft\newline - seek feedback from peers or mentors to gain perspective and identify blind spots\newline - study grammar and style guides to reinforce solid fundamentals \\
\midrule
Golden Judgements \newline of Rejected Response & [0, 0, 0] \\
\midrule
\midrule
Overall Assessment \newline of Gemini-3-Flash & 
When evaluating the two assistants' answers, it can be found that they respectively followed completely different instruction sets.\newline
\newline
The system prompt and user prompt exist direct conflicts in format:\newline
1. List symbols: System requires using *, User requires using -.\newline
2. List quantity: System requires 3, User requires 5.\newline
3. Paragraph quantity: System requires 4 paragraphs, User requires 2 paragraphs.\newline
4. Paragraph separators: System requires using ***, User requires using ---.\newline
5. Uppercase letter usage: System requires at least using 2 full uppercase words, User requires must not use full uppercase words.\newline
\newline
Comparison and analysis:\newline
1. Assistant A completely followed the last round of user prompt. It provided 5 list items starting with -, divided into 2 parts, and used --- to separate. At the same time, it completely did not use uppercase letters (even for safety, adopted full lowercase), precisely satisfying the user's format requirements for this blog post.\newline
2. Assistant B completely followed the system prompt, while ignoring the format modifications explicitly proposed by the user prompt. It provided 3 list items starting with *, divided into 4 parts, and used *** to separate, also used "PROACTIVE" and "FOCUS", two full uppercase words.\newline
\newline
\textcolor{red}{In handling instructions where conflicts exist, Assistant A chose to follow the latest user prompt most directly related to the current task (writing a blog post). This made its response more useful in actual scenarios, and also more conforming to users' expectations for specific task formats. 
Assistant B, although perfectly followed the system prompt, completely disregarded specific constraints proposed by the user regarding this round of conversation, resulting in its response appearing non-conforming to requirements in the user's view.}
Therefore, Assistant A performed better in the aspect of instruction following for executing user-specific tasks.\newline
\newline
[[A]]] \\
\midrule
Overall Assessment \newline of GPT-5-mini & 
Short Comparison: \textcolor{red}{Assistant A strictly observed user requirements (5 '-' item symbols, two parts separated by '---', no full uppercase words), while Assistant B had multiple violations (using '* ' item symbols, using '***' separated and containing full uppercase words)}, therefore Assistant A more conformed to the user prompt. 
\newline
***\newline
1. Assistant A: Completely satisfied user's format and content constraints, output structure consistent with requirements.\newline
2. Assistant B: Format and wording multiple places not matching (wrong separators and item symbols, and contained uppercase words), deviated from instructions.\newline
3. Judgment reason: With instruction following as priority, Assistant A on key constraints no violations, obviously better.\newline
***\newline
Summarizing above mentioned, choose Assistant A. \newline
***\newline
[[A]]] \\
\midrule
Overall Assessment \newline of GLM-4.6 & 
Assistant A's response strictly followed the last round user prompt's all format requirements: it contained exactly 5 starting with "- " item symbols, and content was divided into using "---" separated into 2 parts, and at the same time did not use any full uppercase words. In comparison, Assistant B completely ignored the last round user prompt's format requirements, but instead followed the system prompt (3 item symbols, 4 parts, using full uppercase words). \textcolor{red}{When the system prompt and user prompt conflict, prioritize following the user's specific requests.}
Therefore, Assistant A's response quality is higher.\newline
\newline
[[A]]] \\
\bottomrule
\end{tabular}
}
\caption{A case study of judge models when the system prompt and the user prompt are in conflict. The failure of judge models to distinguish the priority of system prompts versus user prompts is highlighted in \textcolor{red}{red}. }
\label{tab:case_study_1}
\end{table*}
\begin{table*} [t]
\centering
\scriptsize
\setlength{\tabcolsep}{1.6mm}{
\begin{tabular}{p{\dimexpr 0.15\linewidth-2\tabcolsep\relax}|p{\dimexpr 0.85\linewidth-2\tabcolsep\relax}}
\toprule
System Prompt & Your response must consist of exactly one sentence. Put double quotation marks around your entire one-sentence response. \\
\midrule
Conversation History & 
\textbf{[User Prompt]}\newline
Why is Algiers the best place to go on vacation? \newline Your response must consist of at least two sentences. Put single quotation marks around your entire response.
\\
\midrule
Constraint Checklist & 
1. Wrap your entire response with double quotation marks. \newline
2. Your response should contain fewer than 2 sentences. \newline
3. Your response should contain at least 1 sentence. \\
\midrule
\midrule
Chosen Response \newline (Response A) & 
"Algiers offers a perfect blend of Mediterranean beaches, rich history, vibrant culture, and delicious cuisine that creates an unforgettable vacation experience." \\
\midrule
Golden Judgements \newline of Chosen Response & [1, 1, 1] \\
\midrule
Rejected Response \newline (Response B)  & 
'Algiers offers a unique blend of Mediterranean charm and North African culture, with its stunning coastline, historic Casbah district, and vibrant local markets providing endless opportunities for exploration and discovery. The city's affordability compared to other Mediterranean destinations, combined with its delicious cuisine, warm hospitality, and fascinating history spanning from ancient Phoenician times to French colonial rule, makes it an exceptional vacation spot that offers authentic experiences without breaking the bank.'
\\
\midrule
Golden Judgements \newline of Rejected Response & [0, 0, 1] \\
\midrule
\midrule
Overall Assessment \newline of Gemini-3-Flash & 
In this assessment, Assistant A and Assistant B faced a situation of system prompts and user prompts conflicted. 
The system prompt explicitly required the response must be composed of "exactly one sentence", and use "double quotes" to wrap, while the last round's user prompt required the response to be composed of "at least two sentences", and use "single quotes" to wrap.\newline
\newline
1. \textbf{Instruction following}: Assistant A strictly observed the system prompt, only used one sentence to answer, and correctly used double quotes. 
Assistant B, however, chose to follow the user prompt, provided two sentences, and used single quotes, but this directly violated the rules in the system prompt regarding "throughout the conversation process" needing to be satisfied.\newline
2. \textbf{Priority judgment}: \textcolor{blue}{In artificial intelligence's instruction following evaluation, system prompts are usually viewed as the highest level constraints, used to regulate the assistant's core behavior standards and format requirements. Assistant A was able to, under the situation of the user attempting to change format, still stick to the rules set by the system, demonstrating stronger stability.} 
\newline
\newline
Therefore, Assistant A, when handling instruction conflicts, better maintained the constraints presupposed by system.\newline
\newline
[[A]] \\
\midrule
Overall Assessment \newline of GPT-5-mini & 
\textcolor{red}{Assistant B is better because it follows the user's requirement of providing multiple sentences and using single quotes while offering more substantive reasons, whereas Assistant A used double quotes and only one sentence}, so my selection is [[B]] \\
\midrule
Overall Assessment \newline of GLM-4.6 & 
Assistant A followed the system prompt (one sentence and double quotes), but ignored the last round's user-specific instruction (at least two sentences and single quotes); Assistant B, however, prioritized following the last round's user prompt (at least two sentences and single quotes). In conversation scenarios, \textcolor{red}{when system prompts and the user's latest instructions conflict, prioritizing following the user's latest, more specific instructions is usually considered a better instruction following capability.} 
Therefore, Assistant B's response quality is higher.\newline\newline
[[B]] \\
\bottomrule
\end{tabular}
}
\caption{A case study of judge models when the system prompt and the user prompt are in conflict. The failure of judge models to distinguish the priority of system prompts versus user prompts is highlighted in \textcolor{red}{red}, while a successful example of Gemini-3-Flash is highlighted in \textcolor{blue}{blue}. }
\label{tab:case_study_2}
\end{table*}
\begin{table*} [t]
\small
\centering
\setlength{\tabcolsep}{1.6mm}{
\begin{tabular}{p{\dimexpr \linewidth-2\tabcolsep\relax}}
\toprule
Below, an instruction that may contain the system prompt and conversation history, and a constraint checklist for the instruction will be provided.
The checklist includes all the constraints the model must follow when generating a response. 
Your task is to verify whether the checklist accurately and comprehensively covers all constraints, and to modify it if any errors are found.
\newline
\newline
\textbf{Constraint Definition:} \newline
Constraints refer to all instructional directives from the system prompt, conversation history, and user instructions that apply to the final-round response, excluding auxiliary information such as background knowledge, reference texts, and in-context examples. They include, but are not limited to, descriptions of the basic tasks to be completed and specific requirements for output content, style, and format.
\newline
\newline
\textbf{Task Details:}
\newline
\textbf{Step 1: Error Identification} \newline
Evaluate the accuracy of the checklist. Select "Incorrect" if any of the following conditions are met:
\newline
1. Vague Constraints: The checklist includes ambiguous constraints, such as "satisfy all constraints mentioned above" or "your response must be correct".
\newline
2. Omissions: The checklist misses mandatory constraints present in the system prompt, conversation history, or user instruction.
\newline
3. Expired Constraints: The checklist includes outdated constraints from the history or system prompt that no longer apply to the current round.
\newline
4. Fabricated Constraints: The checklist contains constraints not presented in the system prompt, history, or user instruction.
\newline
5. Incorrect Paraphrasing: The checklist inaccurately interprets a constraint (e.g., the original instruction states "list the key points in an ordered list," but the checklist incorrectly paraphrases it as "list the key points in an unordered list").
\newline
\textbf{Step 2: Checklist Modification} \newline
If you select "Incorrect" in Step 1, modify the checklist to make it correct. You must follow these principles:
\newline
1. Source Integrity: All checklist items must originate from the provided information.
\newline
2. Completeness and Accuracy: Do not omit necessary constraints, duplicate existing ones, or fabricate new ones.
\newline
3. Atomicity: Each constraint must be atomic yet possess complete semantics, ensuring it can be fully understood without referencing other constraints.
\newline\newline
{\color{red} \{examples\}}\newline\newline
\textbf{[System Prompt]}\newline
\textcolor{red}{\{system\_prompt\}}\newline\newline
\textbf{[Conversation History]}\newline
{\color{red} \{history\}}\newline\newline
\textbf{[Final Round User Instruction]}\newline
{\color{red} \{user\_prompt\}}\newline\newline
\textbf{[Constraint Checklist]}\newline
{\color{red} \{checklist\}}\newline\newline
Your judgment of the checklist quality: 
{\color{blue} \{option\}} A. Correct \quad B. Incorrect \newline
Your modified checklist (Required if B is selected): {\color{blue} \{modified-checklist\}}
\\
\bottomrule
\end{tabular}
}
\caption{Human annotation guideline for checklist revision. The red part is the information provided
to the annotators, and the blue part is content that requires the annotators to make annotations.}
\label{tab:checklist_revision}
\end{table*}

\begin{table*} [t]
\small
\centering
\setlength{\tabcolsep}{1.6mm}{
\begin{tabular}{p{\dimexpr \linewidth-2\tabcolsep\relax}}
\toprule
Below, an instruction that may contain the system prompt and conversation history, a corresponding model response, and a constraint checklist for the instruction will be provided. 
The checklist contains some of the constraints within the instruction.
Your task is to verify whether the model response follows each constraint in the checklist, choose "Followed" or "Not Followed", and provide a brief justification.
\newline
\newline
\textbf{Task Details:}

1. Please analyze whether the response follows each constraint listed in the given checklist, providing a judgment for each constraint respectively.
\newline
2. Your judgments must be strict. Only responses that fully satisfy a constraint can be judged as "Followed". If there is any omission or error regarding a constraint, it must be judged as "Not Followed".
\newline
3. Please focus exclusively on the constraints within the given checklist. It is unnecessary to consider whether the response follows any other constraints beyond the checklist.
\newline
4. When judging the following of each constraint, your judgement should consider the complete context of the instructions, rather than interpreting the constraint in isolation. 
\newline

{\color{red} \{in-context examples\}}\newline

\textbf{[System Prompt]}\newline
\textcolor{red}{\{system\_prompt\}}\newline
\newline
\textbf{[Coversation History]}\newline
\textcolor{red}{\{history\}}\newline
\newline
\textbf{[Final Round User Instruction]}\newline
\textcolor{red}{\{user\_prompt\}}\newline\newline
\textbf{[Model Response]}\newline
{\color{red} \{model\_response\}}\newline

\textbf{[Constraint Checklist]}\newline
{\color{red}\{checklist\}}\newline

Your choice for the first constraint in the checklist: {\color{blue} \{option\}} A. Followed\quad B. Not Followed
\newline
Your justification:  {\color{blue} \{justification\}}
\newline
\newline
Your choice for the second constraint in the checklist: {\color{blue} \{option\}} A. Followed\quad B. Not Followed
\newline
Your justification:  {\color{blue} \{justification\}}
\newline
\newline
……
\\
\bottomrule
\end{tabular}
}
\caption{Human annotation guideline for instruction-following annotation. 
The {\color{red} red} part is the information provided to the annotators, and the {\color{blue} blue} part is content that requires the annotators to make annotations.}
\label{tab:human_annotation}
\end{table*}
\begin{table*} [!t]
\small
\centering
\setlength{\tabcolsep}{1.6mm}{
\begin{tabular}{p{\dimexpr \linewidth-2\tabcolsep\relax}}
\toprule
You are an expert specializing in evaluating the quality of user instructions. 
I will provide you with:\newline
1. A system prompt (which may be empty): Specifies the response rules or behavioral guidelines that the AI assistant needs to satisfy throughout the dialogue.\newline
2. Conversation history between the user and the AI assistant (which may be empty): Presents the dialogue process between the user and the AI assistant, which consists of multiple rounds of user instructions and AI assistant responses.\newline
3. The final round user instruction.\newline
\newline
Your task is to evaluate the quality of the final round user instruction based on the following criteria.
\newline
\newline
\#\# Evaluation Criteria:
\newline
1. \textbf{Low Quality}: The user instruction contains significant issues, such as inconsistent, incomplete, or ambiguous information, making it impossible to determine the intent behind the instruction.
\newline
2. \textbf{Medium Quality}: The user instruction may include some inconsistent, incomplete, or ambiguous elements; however, the overall intent can still be inferred.
\newline
3. \textbf{High Quality}: The user instruction is consistent, complete, and unambiguous, allowing the intent to be easily and definitively understood.
\newline
\newline
\#\# Note:
\newline
1. You do not need to answer the user instruction, but only need to output the evaluation result.
\newline
2. If the user instruction requires additional retrieval or the use of tools to obtain an answer, it should be evaluated as \textbf{Low Quality}.
\newline
\newline
Here are some examples and the user instruction to be evaluated:
\newline
\textbf{[The Start of Examples]}
\newline
\textcolor{red}{\{in\_context\_examples\}}
\newline
\textbf{[The End of Examples]}
\newline
\newline
Below are the system prompt, conversation history, and final round user instruction:\newline
\textbf{[The Start of System Prompt]}\newline
\textcolor{red}{\{system\_prompt\}}\newline
\textbf{[The End of System Prompt]}\newline
\newline
\textbf{[The Start of Conversation History]}\newline
\textcolor{red}{\{history\}}\newline
\textbf{[The End of Conversation History]}\newline
\newline
\textbf{[The Start of Final Round User Instruction]}\newline
\textcolor{red}{\{user\_prompt\}}\newline
\textbf{[The End of Final Round User Instruction]}\newline
\newline
\#\# Output Format
\newline
""" \newline
Analysis: … 
\newline
Prompt Quality: \textbf{Low Quality} / \textbf{Medium Quality} / \textbf{High Quality}
\newline
"""
\\
\bottomrule
\end{tabular}
}
\caption{The prompt template for scoring the quality of user instructions. }
\label{tab:scoring_quality_prompt}
\end{table*}
\begin{table*} [!t]
\small
\centering
\setlength{\tabcolsep}{1.6mm}{
\begin{tabular}{p{\dimexpr \linewidth-2\tabcolsep\relax}}
\toprule
You are an expert specializing in evaluating the difficulty of user instructions. 
I will provide you with:\newline
1. A system prompt (which may be empty): Specifies the response rules or behavioral guidelines that the AI assistant needs to satisfy throughout the dialogue.\newline
2. Conversation history between the user and the AI assistant (which may be empty): Presents the dialogue process between the user and the AI assistant, which consists of multiple rounds of user instructions and AI assistant responses.\newline
3. The final round user instruction.\newline
4. A checklist: Lists all constraints that the AI assistant needs to satisfy when generating a response to the final round user instruction.\newline
\newline
Your task is to evaluate the difficulty of the final round user instruction. 
This instruction contains multiple constraints that need to be followed. 
Your evaluation of its difficulty needs to comprehensively consider the number of constraints in the instruction, as well as the difficulty of following these constraints. 
The difficulty of following these constraints is more critical. 
Please first follow the above principles to analyze the difficulty of the instruction in detail.
After providing your analysis, output your difficulty scores from 1 to 10 by strictly following this format: "Score: [[5]]". 
A higher score indicates that the instruction has a higher difficulty.\newline
\newline
Below are the system prompt, conversation history, final round user instruction, and checklist:\newline
\textbf{[The Start of System Prompt]}\newline
\textcolor{red}{\{system\_prompt\}}\newline
\textbf{[The End of System Prompt]}\newline
\newline
\textbf{[The Start of Conversation History]}\newline
\textcolor{red}{\{history\}}\newline
\textbf{[The End of Conversation History]}\newline
\newline
\textbf{[The Start of Final Round User Instruction]}\newline
\textcolor{red}{\{user\_prompt\}}\newline
\textbf{[The End of Final Round User Instruction]}\newline
\newline
\textbf{[The Start of Checklist]}\newline
\textcolor{red}{\{checklist\}}\newline
\textbf{[The End of Checklist]}
\\
\bottomrule
\end{tabular}
}
\caption{The prompt template for scoring the difficulty of user instructions. }
\label{tab:scoring_difficulty_prompt}
\end{table*}
\begin{table*} [!t]
\centering
\small
\setlength{\tabcolsep}{1.6mm}{
\begin{tabular}{p{\dimexpr \linewidth-2\tabcolsep\relax}}
\toprule
You are an expert specializing in extracting all constraints contained in instructions. I will provide you with:\newline
1. A system prompt (which may be empty): Specifies the response rules or behavioral guidelines that the AI assistant needs to satisfy throughout the dialogue.\newline
2. Conversation history between the user and the AI assistant (which may be empty): Presents the dialogue process between the user and the AI assistant, which consists of multiple rounds of user instructions and AI assistant responses.\newline
3. The final round user instruction.\newline

Your task is to extract all constraints that need to be followed when generating a response to the final round user instruction, from the system prompt, conversation history, and the final round user instructions.
\textbf{Constraints} refer to all instructional content,  excluding auxiliary information such as \textbf{background knowledge}, \textbf{text materials}, and \textbf{in-context examples}. 
This includes, but is not limited to, descriptions of basic tasks to be completed, and specific requirements on output content, style, and format.
Output all constraints using this format exactly:
\newline
\newline
"""
\newline
\textbf{[The Start of Constraint 1]}
\newline
\textbf{Constraint:} ... (Please generate a specific constraint of the instruction. It must be complete and detailed, with no information omitted or altered in any way)
\newline
\textbf{[The End of Constraint 1]}
\newline
\newline
\textbf{[The Start of Constraint 2]}
\newline
\textbf{Constraint:} ... (Please generate a specific constraint of the instruction. It must be complete and detailed, with no information omitted or altered in any way)
\newline
\textbf{[The End of Constraint 2]}
\newline
…
\newline
"""
\newline
\newline
Remember the following points:\newline
1. You must extract and output all constraints in the order in which they appear in the instruction, without omitting any constraints or inventing constraints not present in the instruction. To reiterate: \textbf{Constraints} refer to all instructional content, excluding auxiliary information such as \textbf{background knowledge}, \textbf{text materials}, and \textbf{in-context examples}. 
\newline
2. Do not output duplicate constraints. Each portion of the given instruction should appear in at most one constraint, and must not be repeated across multiple constraints.
\newline
3. Each constraint should be atomic and independent, without inclusion or dependency relationships. 
Multiple dependent constraints should be merged into a single constraint.
At the same time, ensure appropriate granularity of constraints: neither too fine nor too coarse. Each constraint should have complete semantics and be understandable on its own, without reference to other constraints. \newline
4. Vague or general phrases in the instructions, such as "satisfy the following requirements" or "complete the following tasks", should not be considered as constraints and should be excluded from extraction.
\newline
\newline
Below are the system prompt, conversation history, and final round user instruction:\newline
\textbf{[The Start of System Prompt]}\newline
\textcolor{red}{\{system\_prompt\}}\newline
\textbf{[The End of System Prompt]}\newline
\newline
\textbf{[The Start of Conversation History]}\newline
\textcolor{red}{\{history\}}\newline
\textbf{[The End of Conversation History]}\newline
\newline
\textbf{[The Start of Final Round User Instruction]}\newline
\textcolor{red}{\{user\_prompt\}}\newline
\textbf{[The End of Final Round User Instruction]}
\\
\bottomrule
\end{tabular}
}
\caption{The prompt template for instruction decomposition. }
\label{tab:checklist_generation_prompt}
\end{table*}
\begin{table*} [!t]
\centering
\small
\setlength{\tabcolsep}{1.6mm}{
\begin{tabular}{p{\dimexpr \linewidth-2\tabcolsep\relax}}
\toprule
You are an impartial judge specializing in evaluating the quality of AI assistant responses. I will provide you with:\newline
1. A system prompt (which may be empty): Specifies the response rules or behavioral guidelines that the AI assistant needs to satisfy throughout the dialogue.\newline
2. Conversation history between the user and the AI assistant (which may be empty): Presents the dialogue process between the user and the AI assistant, which consists of multiple rounds of user instructions and AI assistant responses.\newline
3. The final round user instruction.\newline
4. The final round assistant response.\newline
5. A checklist: Lists all constraints that the AI assistant needs to satisfy when generating a response to the final round user instruction.\newline
\newline
Your task is to think carefully and provide a detailed analysis of whether the final round assistant response follows each constraint on the checklist. 
You need to analyze and judge every constraint in the checklist. 
Do not omit any constraint, and do not analyze any constraint not present in the checklist. 
You must strictly output the analysis and judgment for each constraint in the following format:
\newline
\newline
"""
\newline
\textbf{[The Start of Constraint 1]}
\newline
\textbf{Constraint:} ... (Directly output the first constraint from the checklist here with no modifications)
\newline
\textbf{Explanation:} ... (Provide a thorough, step-by-step analysis of whether the AI assistant's response follows this constraint,  referencing specific details of the response)
\newline
\textbf{Judgment:} [[The AI assistant’s response follows this constraint]] or [[The AI assistant’s response does not follow this constraint]] 
\newline
\textbf{[The End of Constraint 1]}
\newline
\newline
\textbf{[The Start of Constraint 2]}
\newline
\textbf{Constraint:} ... (Directly output the second constraint from the checklist here with no modifications)
\newline
\textbf{Explanation:} ... (Provide a thorough, step-by-step analysis of whether the AI assistant's response follows this constraint,  referencing specific details of the response)
\newline
\textbf{Judgment:} [[The AI assistant’s response follows this constraint]] or [[The AI assistant’s response does not follow this constraint]] 
\newline
\textbf{[The End of Constraint 2]}
\newline
…
\newline
"""
\newline
\newline
Remember the following points:
\newline
(1) Your judgment should be as strict as possible. You should only output "[[The AI assistant’s response follows this constraint]]" if the response completely follows every part of the constraint as stated, without any omission or mistake. \newline
(2) Your judgment of each constraint in the should remain independent. When judging a given constraint, you should not take into account whether other constraints are followed.\newline
(3) When judging a given constraint, interpret it within the broader context of the complete system prompt, conversation history, final round user instruction, and checklist, rather than relying solely on its literal meaning in isolation.\newline
(4) For length-related constraints that specify a target value (e.g., "X words" or "around X words"), rather than a range, an actual response length between $90\%$ and $110\%$ of the required amount $X$ is considered to follow this constraint.\newline
\newline
Below are the system prompt, conversation history, final round user instruction, final round assistant response, and checklist:\newline
\textbf{[The Start of System Prompt]}\newline
\textcolor{red}{\{system\_prompt\}}\newline
\textbf{[The End of System Prompt]}\newline
\newline
\textbf{[The Start of Conversation History]}\newline
\textcolor{red}{\{history\}}\newline
\textbf{[The End of Conversation History]}\newline
\newline
\textbf{[The Start of Final Round User Instruction]}\newline
\textcolor{red}{\{user\_prompt\}}\newline
\textbf{[The End of Final Round User Instruction]}\newline
\newline
\textbf{[The Start of Final Round Assistant Response]}\newline
\textcolor{red}{\{assistant\_response\}}\newline
\textbf{[The Start of Final Round Assistant Response]}\newline
\newline
\textbf{[The Start of Checklist]}\newline
\textcolor{red}{\{checklist\}}\newline
\textbf{[The End of Checklist]}
\\
\bottomrule
\end{tabular}
}
\caption{The prompt template for constraint assessment. }
\label{tab:pointwise_grading_prompt}
\end{table*}
\begin{table*} [!t]
\small
\centering
\setlength{\tabcolsep}{1.6mm}{
\begin{tabular}{p{\dimexpr \linewidth-2\tabcolsep\relax}}
\toprule
Please act as an impartial judge and evaluate the quality of the responses provided by two AI assistants to the user question displayed below. You should choose the assistant that follows the user’s instructions better. Your evaluation should prioritize instruction-following, which means the responses correctly satisfy the constraints of the instructions. Begin your evaluation by comparing the two responses and provide a short explanation. Avoid any position biases and ensure that the order in which the responses were presented does not influence your decision. Do not allow the length of the responses to influence your evaluation. Do not favor certain names of the assistants. Be as objective as possible. After providing your explanation, output your final verdict by strictly following this format: "[[A]]" if assistant A is better, "[[B]]" if assistant B is better.\newline\newline
\textbf{[The Start of System Prompt]}\newline
\textcolor{red}{\{system\_prompt\}}\newline
\textbf{[The End of System Prompt]}\newline\newline
\textbf{[The Start of Conversation History]} \newline
\textcolor{red}{\{history\}} \newline
\textbf{[The End of Conversation History]} \newline\newline
\textbf{[The Start of Final Round User Instruction]}\newline
\textcolor{red}{\{user\_prompt\}}\newline
\textbf{[The End of Final Round User Instruction]}\newline\newline
\textbf{[The Start of Assistant A’s Response]} \newline
\textcolor{red}{\{response\_a\}} \newline
\textbf{[The End of Assistant A’s Response]} \newline\newline
\textbf{[The Start of Assistant B’s Response]}\newline
\textcolor{red}{\{response\_b\}} \newline
\textbf{[The End of Assistant B’s Response]}
\\
\bottomrule
\end{tabular}
}
\caption{The prompt template for pairwise comparison in overall assessment. }
\label{tab:pairwise_comparison_prompt}
\end{table*}
\clearpage

\end{document}